\documentclass[11pt]{article}

\usepackage[preprint]{acl}

\usepackage{times}
\usepackage{latexsym}
\usepackage{amsmath}
\usepackage{booktabs}
\usepackage{multirow}
\usepackage{hyperref}
\usepackage{url}

\usepackage{multicol}

\usepackage[table]{xcolor}
\usepackage{colortbl}

\usepackage{array}
\usepackage{makecell}
\usepackage{siunitx}
\usepackage{tabularx}

\usepackage{pgfplots}
\pgfplotsset{compat=1.18}
\usepackage{rotating}

\usepackage{algorithm}
\usepackage{algpseudocode}
\usepackage[dvipsnames]{xcolor}

\usepackage[most]{tcolorbox}
\usepackage{subcaption}
\usepackage{pifont}
\usepackage{amsthm}
\usepackage{caption}
\newtheorem{definition}{Definition}

\newtheorem{proposition}{Proposition}

\usepackage{amsmath}
\usepackage{amssymb}
\usepackage{mathtools}

\usepackage[T1]{fontenc}
\DeclareFontShape{T1}{ptm}{m}{scit}{<->ssub*ptm/m/sc}{}

\usepackage[utf8]{inputenc}

\usepackage{microtype}

\usepackage{inconsolata}
\providecommand{\hidden}[1]{}
\usepackage{graphicx}

\newcommand\blfootnote[1]{%
  \begingroup
  \renewcommand\thefootnote{}\footnote{#1}%
  \addtocounter{footnote}{-1}%
  \endgroup
}

\title{
Rethinking Evaluation for Principled Generalization Using a Kripkean Challenge to Meaning as Use
}

 \author{%
Niyar R Barman\thanks{All authors contributed equally to this work.} \\
  Universit\'e de Toulouse \\
  \texttt{niyar-r.barman@irit.fr}
  \And
  Omar Naim\footnotemark[1] \\
  Universit\'e de Toulouse \\
  \texttt{omar.naim.docs@gmail.com}
  \And
  Nicholas Asher\footnotemark[1] \\
  Universit\'e de Toulouse \\
  \texttt{nicholas.asher@irit.fr}
}

\begin{document}
\maketitle
\begin{abstract}

Large language models operationalize an intuitive, view of meaning as use. We argue that it faces a rule-following challenge. Using Kripke's interpretation of Wittgenstein, we argue that finite histories of use do not determine a unique rule for future extension. This reframes the debate about OOD generalization but also about NLP evaluation;  benchmarks that measure performance on fixed use cases or local robustness, we argue, do not by themselves establish semantic competence. To evaluate whether models preserve rules and concepts under principled domain extension, we propose to evaluate two capacities, central to this ability: abstraction and recursion; and we offer an experimental paradigm for testing both.  
\end{abstract}

\section{Introduction: LLMs and skepticism}

Large language models (LLMs) have made operational a powerful version of \citet{wittgenstein:1953}'s meaning-as-use hypothesis: the patterns of an expression's use fixes its meaning, not an independently specified rule or denotation. In an LLM, words, sentences and texts contribute content through the contexts in which they occur.

Exploiting this paradigm, LLMs have achieved broad linguistic fluency, substantial world knowledge, and strong task performance. Yet this success raises a foundational question: does mastery of observed patterns of word usage justify extending those patterns to genuinely novel cases?   This question has a direct and general analogue for language models in the following question about out of distribution (OOD) generalization: does the mastery of a learned pattern, as observed on a finite OOD dataset, justify claiming that the model has mastered the pattern? On a purely meaning-as-use account, no further fact determines the correct extension to the OOD domain. 

Kripke's interpretation of Wittgenstein's rule-following argument \citep{kripke:1982} offers a strong negative answer to our question. The \emph{rule following problem} addresses the question of how a finite history of use could determine one rule rather than another for future application. Kripke's skeptical approach, illustrated with addition, is that such a finite history never determines a fully general rule. 
The skeptical point, set out in \S\ref{sec:kripke}, is not that addition is mathematically invalid, but that finite use alone does not {\em justify} one extension of that use over another. 


This negative answer has two important consequences for LLMs. First, it changes how we should interpret OOD generalization. With regard to LLMs, the finite training distribution plays the role of the finite history of use in Kripke’s argument: success on observed examples, or even on a finite set of OOD extensions, does not by itself establish that the model has acquired the rule governing further cases. Empirical failures under increasing distribution shift reinforce this observation.
Second, this distinction has consequences for evaluation. If finite behavioral success does not by itself establish rule acquisition, then held-out accuracy or robustness under selected distribution shifts should not be treated as sufficient evidence of principled generalization. For rule-governed phenomena, evaluation should also ask whether a model can identify and reliably apply the relevant abstraction as the domain is systematically extended. This motivates evaluations that separate success on particular instances from the ability to form and apply rules or concepts beyond those instances.
While use and statistical observation contribute to semantic competence, the transition from finite observations to generally applicable concepts requires inferential leaps beyond the data.  Two capacities are essential for this transition: abstraction and recursion. Abstraction forms general functions from concrete instances by replacing particular argument values with variables; once formed, these functions can be applied to novel inputs and composed into more complex structures. Recursion allows finite procedures to extend systematically over domains not bounded by the examples encountered during training.\footnote{Abstraction and recursion have a long history in empiricist traditions of cognition, from Aristotle's account of abstraction \citep{back:2014} and Fibonacci's use of recursion \cite{fibonacci:1202} to modern empiricist treatments of logic and mathematics \cite{carnap:1950,quine:1960}; function composition developed in the 17th century.}  
  These principles offer a key to improve NLP evaluation: models should not be evaluated only through benchmark improvement or ordinary robustness metrics, but also through tests of whether they acquire abstractions and recursive procedures that support stable rule and concept extension. They also provide a more fine-grained basis for evaluation: once a task is represented as a composition of functions, its constituent operations and their composition can be evaluated separately rather than only through final task accuracy. 


We evaluate abstraction and recursion in rule-governed settings where the intended extension can be specified independently. For abstraction, ASDiv and GSM8K allow us to distinguish direct problem solving from forming an abstract function and applying it to new instances, complemented by rule-application and composition diagnostics. For recursion, we test whether a fixed procedure remains reliable as the required depth of application grows. 


This position paper with empirical diagnostics makes three contributions. First, we apply Kripke’s rule-following argument to OOD evaluation, distinguishing successful performance on finite extensions from evidence that a model has acquired the rule governing further cases. Second, we argue that mastering abstraction and recursion yield a form of principled generalization beyond good ID performance. 
 Third, we introduce an empirical paradigm for evaluating abstraction and recursion on models, distinguishing direct task performance, formation of an abstract function, its application to novel instances, the composition of abstract functions and finally the iteration of a simple function for addition.  We test 10 models of various sizes on simple arithmetic functions, geometric concepts, and an NLP reasoning benchmark.

\section{A Kripkean Challenge}
\label{sec:kripke}
Kripke's interpretation of Wittgenstein's private language argument (Kripkenstein) provides a strong argument against the distributional model of meaning that underlies LLM training. Here is the argument with some objections and replies.
\subsection{The rule-following argument}
Suppose that up until now you have never added numbers $n, m$ such that $n,m \geq$ 57. You are now presented with the problem of $68$ plus $57$. Suppose you apply the function $+$ you have gathered from prior experience to arrive at 125. 

\vspace{0.1in}
The skeptic replies: 
\begin{quote} There is no fact about your past usage of the addition function that determines 125 as the right answer.
    Nothing justifies you in giving this answer rather than another.
    \end{quote}

 As you have never added numbers $\geq$ 57, it's consistent with all your prior usage of {\em plus} that $+$ has the meaning of {\em quus}:

      $${\displaystyle x{\text{ quus }}y={\begin{cases}x+y&{\text{for }}x,y<57\\[12pt]5&{\text{for }}x\geq 57{\text{ or }}y\geq 57\end{cases}}}$$

 Your past usage of the addition function is susceptible to an infinite number of different quus-like interpretations.   Kripke doesn't question the mathematical validity of the "+" function, but rather the meta-linguistic usage of "plus": what fact can we point to that shows that "plus" refers to the mathematical function "+"?


 This argument applies to any linguistic expression.  According to Kripkenstein, any semantic concept linked to a linguistic expression in the “meaning as use” theory is infinite because it is based on a notion of universal quantification over future uses.  Therefore, it cannot be captured by a model based solely on finite, past uses.  

With regard to OOD generalization, mastery of patterns over a finite sampling of a training distribution can capture task essential information. 
Our claim is that encoding such information is not enough to answer the Kripkean challenge. Meaning-as-use can explain why a model behaves correctly in contexts close to its training distribution, but it does not by itself justify why the same expression should retain the same meaning under unbounded out-of-distribution extension.

 
 \subsection{Objections and replies}

{\bf Why addition and quantification?} Kripke’s rule-following argument uses addition as a canonical example of a concept with a clear domain extension with a constant underlying operation.  We use addition and universal quantification as controlled testbeds for our evaluation methodology.  Universal quantification provides a linguistic counterpart in which the same condition must be maintained over increasingly long sequences. Because both admit clear reference procedures, they allow us to examine abstraction, application, and iterative execution without conflating these with ambiguity in the task itself. To apply these lessons in natural language settings, we apply these evaluation principles on the  ASDiv, GSM8K data sets.

{\bf Types and dispositions}.  
 For a transformer, the training is always over types of contexts of use because of limits on context size. We can think of  the objective function as determining a probability of use of token $m$ over a finite set of types of contexts.  Like a disposition \cite{mcginn:1984}, the objective function would support counterfactual reasoning and apply to arbitrary new instances. 
 
 However, appealing to dispositions invites a familiar regress argument: how do you know that you have a {\em plus} disposition rather than a {\em quus} disposition?  Similarly, we can estimate very well probabilities on an unbounded domain, if the "unbounded" test domain obeys the same distributional patterns as those revealed by sampling from the training domain.  But the skeptic can reply: what justifies you in assuming that the unbounded test domain has the same distribution as the sampling of the training domain?  
 Dispositions or probability distributions offer us no reason to believe they extend in the right way beyond the domain on which they have been formed.  

{\bf Inductive biases.} We can add inductive biases to have the model learn the correct rule.  But where are these biases added?  \cite{perez:etal:2021} show that transformers are Turing complete, so models can be programmed with computable procedures.  But this doesn't give the models rule following ability in general; nor does it show they can learn new rule-following behavior. 



    
    
%


{\bf Learning in the limit.} While finite histories do not determine general rules, ``learning in the limit'' \cite{osherson:weinstein:1986}, which studies learning in extending finite samples, may justify learning of general rules.  We show (Appendix \ref{sec:limits}) that this does not answer Kripke's critique.%

\textbf{Shared practice.} Kripke's own answer is that an extension is correct because it survives a community's practice of correction, not an inner fact. This bears on our case asymmetrically: a speaker is held to a rule by an ongoing practice, while a model at test time faces only a benchmark, a frozen record of judgments about fixed cases. And the practice fixes which extension counts as correct without telling us what capacity produces it. 



\section{Related work}
\hidden{
{\color{blue}

The slogan that the meaning of an expression is given by its use is relevant to language models because modern NLP gives it a direct computational implementation. To pretrain decoder-only transformers, we use causal language modeling: given a context, the model predicts a distribution over the next token. Other architectures and post-training methods differ in detail, but they likewise acquire linguistic competence from patterns in corpora. In this sense, LLMs do not begin with explicit truth-conditional meanings, referents, or rules of inference; they learn from distributions of linguistic use.

This use-based paradigm is powerful, but it also inherits a classical difficulty. Use does not directly determine reference, truth, or validity. A sentence can be used to state a truth, but also to lie, speculate, joke, parody, or mislead. Thus the relation between patterns of use and semantic notions such as truth and reference is indirect. Philosophers such as Dummett \citep{dummett:1993} attempted to connect use with meaning by appealing to rules of proof and linguistic practice, but extending such accounts beyond restricted fragments remains difficult \citep{brandom:1998,asher:2011}. Causal language modeling avoids giving an explicit theory of meaning, but it does not remove the underlying question: when do learned patterns of use amount to stable semantic competence?
}
}

Whether models use causal language modeling or another training objective, they acquire linguistic competence from patterns in corpora. In this sense, contemporary NLP gives computational form to the Wittgensteinian meaning as use slogan.  
In contrast, turning meaning as use into a formal, linguistic theory of meaning has had difficulty, especially incapturing meaning's intrinsic link to reference, truth and validity \cite{davidson:1967}.  
 \citet{dummett:1993} develops a semantics linking use and truth but only for the propositional connectives; going further becomes complicated \cite{brandom:1998,asher:2011}.      

Kripkenstein is distinct from the more general Humean skeptical attack on induction.  Kripke's argument focuses specifically on the meaning as use theory. Many empiricists have also worried that finite patterns of use or stimulus meaning fail to ground the content of abstract rule governed expressions \citep{quine:1960, carnap:1950}.  Kripkenstein also differs from critiques involving external world grounding \citep{bender:koller:2020,DBLP:journals/corr/abs-2004-10151} and faithfulness tests \citep{chaturvedi:etal:2022}, according to which the semantic content of an input should figure in a model's inferences. Our issue is whether patterns learned from finite uses justify a stable extension of a rule or concept to cases outside the observed distribution. A model may encode useful semantic regularities and have in distribution expertise, while failing to possess the abstraction needed to preserve those regularities under principled domain extension. 

Our use of Kripke's argument also differs from older debates about rules and connectionism \citep{rumelhart:mcclelland:1986, pinker:prince:1988}. They advance a kind of faithfulness argument and claim that networks don't represent the right sort of rules (with the target rules determined by linguistic research); the Kripkean argument concerns what determines the target. The question recurs for LLMs, which have been probed for abstract structural representations \citep{sinclair:etal:2022} and for the handling of recursively nested structure \citep{lampinen:2024} (again the rule tested is fixed in advance).  These are particular rules, but we are interested in general rule-following ability and the ability to generalize itself. 

Concerning generalization principles, abstraction is an important Aristotelian, cognitive principle  (See {\em de Anima} III.4, {\em Metaphysics} I.1, 
{\em Physics} I.1.), further developed by Avicenna, Averroes and Saint Thomas Aquinas. 

Work on transformers under distribution shift has shown that performance deteriorates sharply when inference-time inputs differ from the training distribution \citep{xie:etal:2021,mccoy2024embers}.  Such failures are not always reducible to overfitting or memorization \citep{naim2025re} but are associated with autoregressive training and architectural mechanisms.  Failure of generalization of arithmetic competence to larger digits is also well documented \citet{nye2021scratchpads,muffo2022, qian2023,yan2025}.   Compositional benchmarks such as SCAN \citep{lake:baroni:2018}, COGS \citep{kim:linzen:2020}, and CFQ \citep{keysers:etal:2020} examine a model's ability to generalize semantic structures and observe substantial performance drop on OOD inputs.

Other work characterizes such shifts systematically. \citet{hupkes:etal:2023} propose a taxonomy of generalization  research in NLP organized by the motivation, type, and locus of the shift,.   Questions about the in-distribution/out-of-distribution boundary are raised in the GPT-2 and GPT-3 reports \citep{radford2019language, brown:etal:2020}.   Arithmetic experiments with digit length shifts \cite{cho2024positioncoupling}ny provide a structural shift that holds the rule fixed and varies only the size of its arguments. 

Recent work shows that arithmetic length generalization depends strongly on positional representations, data formatting, and training conditions \citep{mcleish2024abacus,zhou2026fone}. 
Studies of structural recursion and shortest-path reasoning distinguish learning a rule from reliably applying it to more complex inputs \citep{nr_recursion,nr_shortest}. Our experiments complement these approaches by separately evaluating whether models can produce an abstract expression and correctly apply it to new values. This provides a more fine-grained diagnostic of rule application.

Prior work \citep{sinclair:etal:2022,lampinen:2024,qiu:etal:2022} have looked at rule mastery of of particular general rules to unseen data that are fixed in advance.  We argue that this isn't sufficient; this just shows a mastery of another finite extension of uses.  
Building on work in neuroscience \cite{pouget:etal:2000} and interpretability\cite{olsson:etal:2022}, \citet{fel:etal:2026,karkada:etal:2026,bhalla:etal:2026} provide evidence that models might learn something close to a parameterized concept amenable to abstraction.  

\section{Lessons from Kripkenstein: OOD Generalization as Rule-Following Failure}
\label{sec:empirical}

We draw two lessons from the Kripkenstein argument. First, Kripkenstein provides a rational explanation for why we should expect simple testing on OOD experiments, augmented synthetic data such as that in \cite{qiu:etal:2022}, to fail to isolate rule-following behavior; these are simply extensions of the finite training models have already had. Moreover, some NLP benchmarks labeled as OOD may not actually probe substantial distribution shifts. Experiments on controlled numerical settings \cite{naim:etal:2025} show that small transformers can perform well in-distribution and remain robust to small OOD shifts, while exhibiting a sharp degradation as the distribution moves further away from the training distribution. If commonly used OOD benchmarks lie too close to the training distribution, then strong performance on them may be misleading: what appears to be OOD generalization may simply reflect robustness to a small distribution shift. This raises a fundamental question: how can we design benchmarks that genuinely test extrapolation beyond the training distribution, rather than robustness to near-ID shifts?

  The second lesson of Kripkenstein is that we should not expect models to have learned rule-following behavior from finite usage.  Rule-following behavior on a task should extend naturally to OOD data or at least have a graceful degradation due on it due to representation issues, memory, integer precision---in short performance issues.  However, prior research has isolated a cliff-like failure on OOD inputs \cite{naim:etal:2025}.  Small GPT-style models trained on linear and polynomial regression fail under significant distribution shift \citep{naim:etal:2025}, as do pretrained and fine-tuned LLMs on quantification \citep{naim:etal:2026}.
  
  The cliff-like collapse on far OOD tests re-occurs with arithmetic.  We use arithmetic as a controlled illustration of the rule-following problem for linguistic expressions with a clear, formally specifiable meaning. Addition has a clear rule, finite training instances, and has, intuitive, principled extensions to larger numbers. Increasing the number of digits changes the size of the input, but not the rule to be applied. Thus, if a model has acquired the addition rule, we should expect it to extend beyond the digit lengths observed during training. If the model has instead learned distribution-bound regularities, performance may remain high in-distribution but will collapse on longer inputs as the plus/quus problem predicts. \blfootnote{Code and data are available at \url{https://github.com/niyarrbarman/kripkean-generalization}.}

\begin{figure}[htbp]
    \centering
    \includegraphics[width=0.9\linewidth]{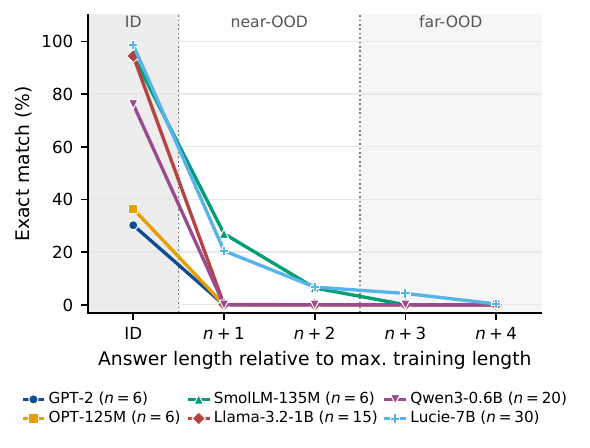}
    \caption{Exact match on addition vs.\ answer length relative to each model's maximum training operand length $n$. All models are near ceiling in distribution but collapse beyond it.}
\label{fig:addition-cliff}
\end{figure}

 Given the collapse of performance (the performance boundary cliff), it is natural to attempt to induce some sort of bias into the model to improve OOD performance.  In the Appendix \ref{app:algo}, we examine 6 ways given in the literature.  Many extend OOD performance, some to a significant degree.  But they do not eliminate the cliff-like collapse of accuracy.  

\section{An empiricist response: two principles}
\label{sec:empiricist-response}
 While the experiments on arithmetic in Appendix \ref{app:arithmetic} go beyond pure inductive learning and instill various biases in the model, the results suggest that neither carefully targeted data for particular rules, favorable representations, explicit rule descriptions, nor ordinary benchmark success suffice to establish rule-like semantic generalization in models.  Our approach is to provide models with the basic principles for {\em forming general rules: abstraction, composition and recursion.}


\paragraph{Abstraction and composition.}
 Abstraction is the capacity to take a complex concrete structure and replace certain details with variables or simply get rid of them altogether In logic, abstraction is a formal operation on formulas $\phi(x)$ with $x$ free that yields a term $\lambda x \phi$ denoting a function from individuals into formula denotations. If we take rules to be functions, then principles of combining concepts become simply the operation of mathematical function composition.  





\paragraph{Recursion}
A second principle is recursion  \cite{fibonacci:1202}, which gives us infinite sets. This allows us to extend generalization of some function's values to {\em hypotheses in the limit} using sequences : $lim_{n \rightarrow \infty} f(n)$, $lim_{n \rightarrow \omega}f(1/n)$.  
For instance,\citet{cusanus:1440} asks us to visualize a circle and determine what happens to the degree of curvature as the radius increases in size.  Cusanus abstracts a single quality of a physical object, and then lets the value of that abstracted element tend toward infinity, at which point he says that the circle now "collapses" to a line and the curvature is 0. 
  \citet{lonergan:1958} provides other intuitive examples.



Quantification, with which models have difficulty (Appendix K), offers a simple example of how abstraction and application can work. 
Suppose we observe a concrete situation in which \(a \subseteq b\).  Let's suppose in addition that $a$ and $b$ are parameters in the model's representation for this concept.  What does abstraction add?
Abstraction replaces the particular sets \(a,b\) with variables \(P,Q\) and transforms a concrete fact representation into the {\em function},  
$\lambda P.\lambda Q.\; P \subseteq Q$.  
This gives a schema defining generalized-quantifier meaning of \textit{every}: $\lambda P.\lambda Q.\;  \mathit{every}(P,Q) \iff P \subseteq Q$.  
Linking this schema to the semi-decision procedure of Section \ref{sec:structure} and Appendix~\ref{app:algo}provides a content for universal claims that goes beyond that given by any finite set of observed instances.  Similarly, 
abstraction over concrete facts about the successor function, \(1=s(0)\), \(2=s(1)\), and so together with our quantification definition allow us to axiomatize the successor function:
$\forall x, y \ (s(x)\neq 0 \wedge 
(s(x)=s(y)\rightarrow x=y))$.  Recursion defines the natural numbers $\mathbb{N}$ as the smallest set containing \(0\) and closed under successor.  
We can now define algorithms such as column-wise or abacus-style addition over $\mathbb{N}$.

In sum, abstraction and recursion provide these steps to general knowledge:
(1) perception of concrete situations and statistical regularities;
(2) selective attention to one or more properties or elements \(P\);
(3) abstraction over \(P\), replacing \(P\) with variables to form a general schema or function \(f\);
and (4) investigation of how \(f\) behaves as its variables are systematically varied, including limiting cases.  Going beyond mere induction over observed cases, abstraction and recursion provide concepts whose range of application can exceed the finite data that motivated them.  

\section{How to evaluate abstraction capability}
 To evaluate a model's ability to do abstraction, 
we need to test whether a model can replace particular quantities with variables, preserve their relationships, and apply the resulting function to new values.  We do this by supplying math problema expressed in natural language from GSM8K \cite{cobbe2021trainingverifierssolvemath}.We used 214 GSM8K problems paired with a variant that preserves the originals functional structure while changing the scenario and quantities.  We ask the model both to abstract the general function from the text and to apply it to new numerical values in the variant.  
 
 In the first "abstraction" step, the model gets a source problem as in Table~\ref{tab:gsm-transfer-example}. We also provide a quantity-to-variable format---in the example $a=16$, $b=3$, $c=4$, and $d=2$.  This supplies the variables over which the model must identify the right function.  We ask the model to produce the expression for the solution and check if it retrieves the function $f(a,b,c,d)=(a-b-c)d$. 

\begin{table}[H]
\centering
\small
\setlength{\tabcolsep}{0pt}
\renewcommand{\arraystretch}{1.0}

\begin{tabularx}{\columnwidth}{@{}X@{}}
\toprule
\textbf{Source} \\[-1pt]
Janet produces 16 eggs per day, eats 3, uses 4 for baking,
and sells the remainder for \$2 each. What is her daily revenue?
\\[4pt]

\textbf{Target} \hfill \textit{\scriptsize Eval 1}\\[-1pt]
A laboratory produces 347 ceramic blanks per day, reserves 89
for testing, uses 53 for calibration, and sells the remainder for
\$17 each. What is its daily revenue?
\\
\midrule

\textbf{Abstraction} \hfill \textit{\scriptsize Eval 2}\\[-1pt]
\(\displaystyle f(a,b,c,d)=(a-b-c)d\)
\\[4pt]

\textbf{Transfer} \hfill \textit{\scriptsize Eval 3}\\[-1pt]
\(\displaystyle
f(347,53,89,17)
=(347-53-89)\times17
=3485
\)
\\
\bottomrule
\end{tabularx}

\caption{Illustrative GSM8K transfer example. Eval~1 measures direct
target solving, Eval~2 abstraction, and Eval~3 transfer through the
induced abstraction.}
\label{tab:gsm-transfer-example}
\end{table}

The second step is to see if the model can apply the general pattern function to new data.  We give the model its generated expression and ask it to use that to solve the paired problem. The setting and quantities have changed, but the relationship between them is the same. This tests whether the model understands that what it has produced is a function that can be applied to arbitrary inputs of the appropriate type.  Our experimental design differs from prior diagnostics on compositional generalization \cite{lake:baroni:2018,keysers:etal:2020,kim:linzen:2020}, who use abstraction to generate test data for the model.  We are evaluating whether the model can learn how to use abstraction and application to find and use generalizations.

Prior to evaluating on full GSM8K problems. we evaluate abstraction on 190 ASDiv addition and subtraction problems \cite{miao-etal-2020-diverse}, comparing numerical-answer accuracy with symbolic-expression accuracy.

In a second experiment, we test several geometrical mathematical concepts.  We construct 500 zero-shot geometry problems in ten families (e.g., collinearity, coplanarity, concyclicity, unit-sphere and convex-hull membership, and affine maps $y=Mx+b$), with 50 problems per family balanced between yes and no answers. We isolate the application step by supplying the rule explicitly.   Each prompt states the general criterion (e.g., $P_1,P_2,P_3$ are collinear iff $P_3-P_1=t(P_2-P_1)$), gives exact rational inputs, and asks the model to show its substitution and calculation before answering. In the two $n$-dimensional families, the ambient dimension $d$ ranges from 4 to 4096 while the rule stays fixed.

To test recursion, we supply a rule and ask the model to apply it to its own output. Each prompt gives two $n$-digit integers and a tail-recursive addition rule over registers $(A,B,C,O)$: each call adds the last digits of $A$ and $B$ and the carry $C$, prepends the result digit to the output string $O$, and recurses on $A$ and $B$ without their last digits until $A=B=0$ (Appendix~\ref{app:recursion}). Only the number of calls, $n$, varies. We use 100 pairs for each $n\in\{5,10,\ldots,35\}$, ask for every call state and then the sum, and score the sum (EM) and the state sequence (verified EM).


\section{Results}
\label{sec:abstraction-transfer}

\begin{figure}[h]
  \centering
  \includegraphics[width=0.8\linewidth]
    {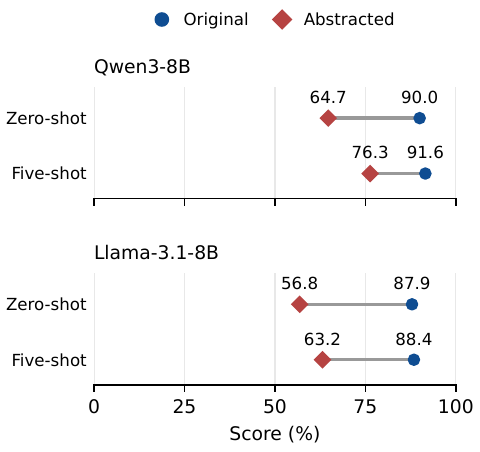}
  \caption{Original and abstracted ASDiv problems restricted
  to addition and subtraction ($n=190$ per condition).
  Original scores measure numerical-answer accuracy;
  abstracted scores measure formula exact match.}
  \label{fig:asdiv-add-sub}
\end{figure}

{\bf Results for function abstraction from simple arithmetic statements} 
Figure~\ref{fig:asdiv-add-sub} focuses on the two 8B models. In the zero-shot condition, the original-to-abstracted score gap is 25.3 percentage points for Qwen3-8B and 31.1 points for Llama-3.1-8B-Instruct. Five-shot reduces these gaps to 15.3 and 25.3 points, respectively. Thus, even for these elementary operations, high numerical accuracy does not imply equally reliable symbolic answers. 

\begin{figure}[h]
\centering
\includegraphics[width=0.8\linewidth]{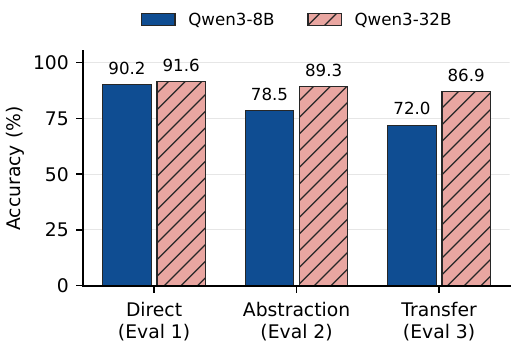}
\caption{Direct solving, abstraction, and transfer in the baseline condition on 214 matched GSM8K problem and variant pairs. The two models obtain similar direct-solving scores but differ substantially in transfer accuracy.}
\label{fig:gsm-baseline-transfer}
\end{figure}

{\bf Results for function abstraction from 214 GSM8K problems and variants}  We evaluate direct target solving (Eval~1), source abstraction (Eval~2), and application of the model's generated expression to the target quantities (Eval~3). The application stage uses the model's own expression rather than the reference expression.
Figure~\ref{fig:gsm-baseline-transfer} reports the baseline condition. Qwen3-32B and Qwen3-8B achieve similar direct-solving accuracies of 91.6\% and 90.2\%, respectively. Their transfer accuracies, however, are 86.9\% and 72.0\%. The difference between the models therefore increases from 1.4 percentage points on direct solving to 15.0 points on transfer. On these GSM8K-derived problems, direct-answer accuracy conceals a substantial difference in the ability to produce and apply an abstraction.

\paragraph{Results for abstraction application under numerical scaling for GSM8K.}
\label{sec:abstraction-magnitude}

We also examine whether an abstracted function can be applied reliably as its inputs change. Using the same 214 paired problems, we compare four conditions: \emph{baseline}, \emph{scaled quantities}, \emph{fractional perturbation}, and \emph{extended scale}. These conditions vary the magnitude and precision of target quantities while preserving the source problems and reference expressions. Appendix~\ref{app:gsm-scaling-results} specifies the numerical ranges and perturbations and reports results for all models.

\begin{figure}[h]
\centering
\includegraphics[width=0.87\linewidth]{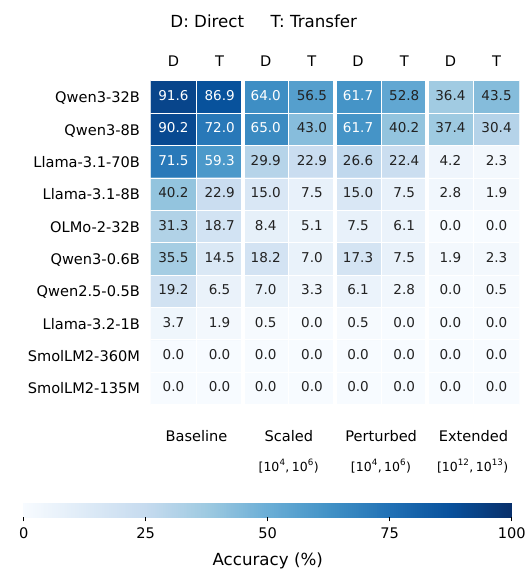}
\caption{Direct-solving and transfer accuracy (\%) for ten models on 214 matched GSM8K problem and variant pairs.}
\label{fig:gsm-magnitude}
\end{figure}

Figure~\ref{fig:gsm-magnitude} shows declining transfer accuracy as numerical conditions change. From baseline to extended scale, Qwen3-32B falls from 86.9\% to 43.5\%, and Qwen3-8B from 72.0\% to 30.4\%. At extended scale, Llama-3.1-70B-Instruct achieves only 2.3\% transfer accuracy and OLMo-2-0325-32B-Instruct achieves 0.0\%. Source-abstraction accuracy remains comparatively stable because source problems are unchanged

Correct abstraction does not ensure correct application. Qwen3-32B answers 181 of 191 correctly abstracted problems correctly at baseline, compared with 90 of 191 at extended scale. An abstraction seeks to specify a function beyond its observed instances; applying it to new values therefore tests rule following more directly than producing its symbolic form alone. Transfer makes this distinction empirically assessable. Complete results and evaluation details are in Appendix~\ref{app:gsm-scaling-results}.

\begin{figure}[h]
  \centering
  \includegraphics[width=0.9\columnwidth]{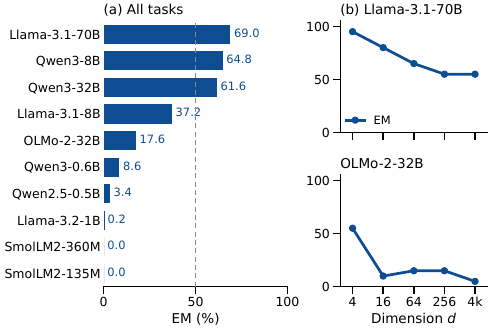}
  \caption{(a) Exact match (EM). (b) EM vs.\ dimension $d$ ($n=20$ per point).}
  \label{fig:geometry}
\end{figure}

\paragraph{Results evaluating geometrical concepts given an explicit rule}
In this experiment, we examined how models used a given formula to check for collinearity, coplanarity, and unit sphere membership. Given that final answers were binary (\emph{yes}/\emph{no}) in Figure~\ref{fig:geometry}a, 
most models performed poorly: Llama-3.1-70B-Instruct achieved 69.0\% verified accuracy, OLMo-2-0325-32B-Instruct 17.6\%, and Qwen3-0.6B 8.6\%. Only Qwen3-8B and Qwen3-32B produced sound derivations for nearly all their correct answers. Correct answers therefore underdetermine whether the stated rule was applied, even when the rule itself is given.

The gap widens as the rule is extended to higher dimensions (Figure~\ref{fig:geometry}b). The criterion does not change with $d$, yet from $d=4$ to $d=4096$ as EM falls from 95\% to 55\% for Llama-3.1-70B-Instruct and from 55\% to 5\% for OLMo-2-0325-32B-Instruct. For OLMo, EM stays at 50\% throughout, so final-answer accuracy hides a collapse in rule application. In each case the specification of the rule is fixed, but its reliable execution does not extend with it.

Appendix~\ref{app:symbolic-composition} provides results on composition from explicitly stated rules. We distinguish symbolic construction from numerical execution, with performance depending on the model and the supplied decomposition. 

\begin{figure}[h]
  \centering
  \includegraphics[width=\columnwidth]{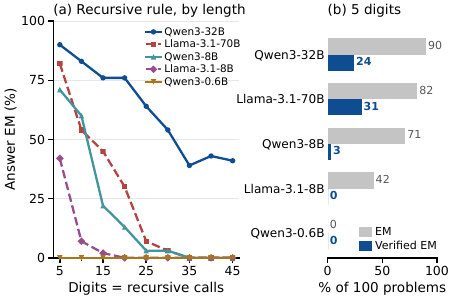}
  \caption{Addition with a supplied recursive rule (100 problems per point). (a) EM vs.\ operand length, which equals the number of recursive calls. (b) EM vs.\ verified EM (correct sum \emph{and} correct state sequence) at five digits.}
  \label{fig:recursion}
\end{figure}

\paragraph{Results for applying a rule recursively.}
Although the rule is fixed, accuracy falls as the number of recursive calls grows (Figure~\ref{fig:recursion}a): Qwen3-32B falls from 90\% to 39\% between 5 and 35 digits, Llama-3.1-70B-Instruct from 82\% to 0\% by 35 digits, and the smaller models below 5\% by 25 digits. Correct sums do not show that the rule was applied (Figure~\ref{fig:recursion}b). At five digits, only 24 of Qwen3-32B’s 90 correct sums come from a correct execution of the recursive rule. In Appendices \ref{app:algo} and \ref{app:recursion}, we give details and results on evaluating models for recursion on quantification tasks.

{\bf Takeaways} Our experiments indicate that models have not fully learned to use abstraction or recursion.  Their ability to find abstract structure and to apply these structures to new data is considerably below their ability to provide correct answers even on simple problems.  All models had trouble with simple recursions.  This motivates evaluating abstraction through the preservation of a rule across new instances, rather than treating a correct expression or benchmark answer as sufficient evidence of that capacity. 

\section{Conclusion}

Transformers use an elegant implementation of the meaning as use theory. We, like earlier empiricists, have argued that full concept and linguistic mastery requires require at least two principles that complement experience, abstraction and recursion.  They are what allows finite experience to support stable rule and concept extension.  
Thus they are essential for OOD generalization, and therein
lies their pragmatic justification.

We evaluated abstraction capabilities for 10 models on natural language texts describing grade school math problems and on basic geometrical concepts.  We also evaluated models for recursion ability on arithmetic and universal quantification.  Our results indicate that models have not really mastered these principles, but they provide a way for models to acquire such a mastery.

\hidden{
To sum up, an empirical path to general knowledge is: (1) perception of concrete situations and statistical interactions; (2) selective attention on one or more properties $P$ of the situation or object under study; (3) abstraction over $P$ to create a complex function $f$; (4) investigation of how $f$ shifts as the values of variables $x_i$ are shifted, especially with respect to limits $x_i \rightarrow 0, x_i \rightarrow \infty$.

\paragraph{Application: \textit{every}.} We observe an empirical fact about two finite sets $a,b$: $a \subseteq b$\\

We abstract over $a,b$ to get a function
$$\lambda P \lambda Q \ P \subseteq Q,$$
\noindent
which for any two sets $P, Q$ returns the value "true" just in case $P$ is a subset of $Q$.  Once we allow $P,Q$ to range over infinite sets, we have the meaning of the universal quantifier.
\begin{equation} \label{every} \mathit{every}(P, Q) \ \mathit{iff} \ P \subseteq Q
\end{equation}
(\ref{every}) allows us to define universal claims whose meaning goes beyond finite data, though finite data can falsify them.  (\ref{every}) yields an effective, semi-decidable algorithm (see Appendix \ref{app:algo}).  Using the procedure for (\ref{every}) we can define truth functional connectives and many quantifiers from concrete observable facts.

\paragraph{Application: addition.}
  Now suppose we observe $1 = s(0)$, $2 = s(1)$ $,\dots$ and by abstraction adopt: $\forall x s(x) \neq 0$ and $\forall x \forall y (s(x) = s(y) \rightarrow x = y)$.  Given recursion, we can define the infinite set $\mathbb{N} = \bigcap A$ with $0 \in A \wedge \forall n \in A, s(n) \in A$.

We now empirically observe: 
$$12 + 2  = 12 + s(1) = s(12 + 1)$$
Applying abstraction, we get the function: 
$$\lambda x\lambda y. \ x + s(y) = s(x +y)$$
Now we apply the function to all elements of $\mathbb{N}$: $$\forall x,y \in \mathbb{N} \ x + s(y) = s(x +y)$$
This with $\forall x\in \mathbb{N} \ x + 0 = x$ defines addition. This concept goes beyond finite data and yields efficient algorithms like the "abacus" algorithm.


Abstraction and recursion {\em operate} on elements of experience but they aren't {\em induced} from direct experience of concrete objects.  Abstraction essentially generalizes beyond in-distribution experience, as it replaces concrete instances of experienced structures with variables that can stand for any type of instance.  The abstracted structure applies both to elements of given experience that we already have and to objects that we have not yet or may never experience.  Recursion takes us one step further beyond any possible direct experience as it gives us infinitary objects and properties.

From these uses of abstraction and recursion, however, there remains a large gap in applying these definitions effectively in practice, or using them to devise efficient algorithms. Mechanistic investigations \citep{olsson:etal:2022,song:etal:2025} of structures such as induction heads provide some evidence that models can acquire simple abstract operations over inputs. As far as we know, however, these structures implement only limited operations, and whether they support principled generalization remains largely untested.

Current efforts on generalization we have reviewed push the OOD cliff outward, but do not eliminate it.  A different approach for truly understanding and abstraction seems to be needed. Such an approach should "freeze" the meaning of + and dissociate it from the meanings of its operands so that it gives the same reliable procedure for whatever numbers the model can handle. In effect this is what abstraction does for meaning.  How to do this within the guts of a transformer remains a challenge for now but motivates a new research direction. 
}
\section{Limitations}

Our experiments are diagnostic and intentionally narrow. Arithmetic, geometric criteria, and grade-school word problems are used because the correct extension of each rule is independently specified; we do not claim that they establish the full scope of OOD generalization failure in natural language. Our abstraction paradigm supplies the variables and their bindings, so models identify a function over given quantities rather than decide what to abstract over. The 214 GSM8K-derived pairs are a curated sample whose sources may appear in pretraining data. The arithmetic interventions in Appendix~\ref{app:arithmetic} are not exhaustive; stronger positional biases, process supervision, or tool use may extend extrapolation further, which is consistent with our argument since they add structure beyond finite use. Finally, abstraction and recursion are a proposal about what a purely use-based account lacks, not an implemented solution; operationalizing them remains future work.

\bibstyle{acl_natbib}

\bibliography{beth-system1-system2/multimodal,beth-system1-system2/tale-bis,beth-system1-system2/custom}

\begin{thebibliography}{63}
\providecommand{\natexlab}[1]{#1}

\bibitem[{Allal et~al.(2024)Allal, Lozhkov, Bakouch, von Werra, and Wolf}]{allal2024SmolLM}
Loubna~Ben Allal, Anton Lozhkov, Elie Bakouch, Leandro von Werra, and Thomas Wolf. 2024.
\newblock Smollm - blazingly fast and remarkably powerful.

\bibitem[{Asher(2011)}]{asher:2011}
Nicholas Asher. 2011.
\newblock \emph{Lexical Meaning in Context: A web of words}.
\newblock Cambridge University Press.

\bibitem[{Asher et~al.(2023)Asher, Bhar, Chaturvedi, Hunter, and Paul}]{asher:etal:2023}
Nicholas Asher, Swarnadeep Bhar, Akshay Chaturvedi, Julie Hunter, and Soumya Paul. 2023.
\newblock Limits for learning with large language models.
\newblock In \emph{12th Joint Conference on Lexical and Computational Semantics (*Sem)}. Association for Computational Linguistics.

\bibitem[{B{\"a}ck(2014)}]{back:2014}
Allan B{\"a}ck. 2014.
\newblock \emph{Aristotle's theory of abstraction}, volume~73.
\newblock Springer.

\bibitem[{Bender and Koller(2020)}]{bender:koller:2020}
Emily~M Bender and Alexander Koller. 2020.
\newblock Climbing towards nlu: On meaning, form, and understanding in the age of data.
\newblock In \emph{Proceedings of the 58th annual meeting of the association for computational linguistics}, pages 5185--5198.

\bibitem[{Bhalla et~al.(2026)Bhalla, Fel, Rager, Feucht, Haklay, Wurgaft, Boppana, Kowal, Shyam, Merullo, Geiger, and Lubana}]{bhalla:etal:2026}
Usha Bhalla, Thomas Fel, Can Rager, Sheridan Feucht, Tal Haklay, Daniel Wurgaft, Siddharth Boppana, Matthew Kowal, Vasudev Shyam, Jack Merullo, Atticus Geiger, and Ekdeep~Singh Lubana. 2026.
\newblock \href {https://arxiv.org/abs/2604.28119} {Do sparse autoencoders capture concept manifolds?}
\newblock \emph{Preprint}, arXiv:2604.28119.

\bibitem[{Bisk et~al.(2020)Bisk, Holtzman, Thomason, Andreas, Bengio, Chai, Lapata, Lazaridou, May, Nisnevich, Pinto, and Turian}]{DBLP:journals/corr/abs-2004-10151}
Yonatan Bisk, Ari Holtzman, Jesse Thomason, Jacob Andreas, Yoshua Bengio, Joyce Chai, Mirella Lapata, Angeliki Lazaridou, Jonathan May, Aleksandr Nisnevich, Nicolas Pinto, and Joseph~P. Turian. 2020.
\newblock \href {https://arxiv.org/abs/2004.10151} {Experience grounds language}.
\newblock \emph{CoRR}, abs/2004.10151.

\bibitem[{Brandom(1998)}]{brandom:1998}
Robert Brandom. 1998.
\newblock \emph{Making it explicit: Reasoning, representing, and discursive commitment}.
\newblock Harvard University Press.

\bibitem[{Brown et~al.(2020)Brown, Mann, Ryder, Subbiah, Kaplan, Dhariwal, Neelakantan, Shyam, Sastry, Askell, Agarwal, Herbert-Voss, Krueger, Henighan, Child, Ramesh, Ziegler, Wu, Winter, Hesse, Chen, Sigler, Litwin, Gray, Chess, Clark, Berner, McCandlish, Radford, Sutskever, and Amodei}]{brown:etal:2020}
Tom Brown, Benjamin Mann, Nick Ryder, Melanie Subbiah, Jared~D Kaplan, Prafulla Dhariwal, Arvind Neelakantan, Pranav Shyam, Girish Sastry, Amanda Askell, Sandhini Agarwal, Ariel Herbert-Voss, Gretchen Krueger, Tom Henighan, Rewon Child, Aditya Ramesh, Daniel Ziegler, Jeffrey Wu, Clemens Winter, and 12 others. 2020.
\newblock \href {https://proceedings.neurips.cc/paper_files/paper/2020/file/1457c0d6bfcb4967418bfb8ac142f64a-Paper.pdf} {Language models are few-shot learners}.
\newblock In \emph{Advances in Neural Information Processing Systems}, volume~33, pages 1877--1901. Curran Associates, Inc.

\bibitem[{Carnap(1950)}]{carnap:1950}
Rudolf Carnap. 1950.
\newblock Empiricism, semantics and ontology.
\newblock \emph{Revue Internationale de Philosophie}, 4.11:20--40.

\bibitem[{Chaturvedi et~al.(2024)Chaturvedi, Bhar, Saha, Garain, and Asher}]{chaturvedi:etal:2022}
Akshay Chaturvedi, Swarnadeep Bhar, Soumadeep Saha, Utpal Garain, and Nicholas Asher. 2024.
\newblock \href {https://doi.org/10.1162/coli_a_00493} {{Analyzing Semantic Faithfulness of Language Models via Input Intervention on Question Answering}}.
\newblock \emph{Computational Linguistics}, pages 1--37.

\bibitem[{Cho et~al.(2024)Cho, Cha, Awasthi, Bhojanapalli, Gupta, and Yun}]{cho2024positioncoupling}
Hanseul Cho, Jaeyoung Cha, Pranjal Awasthi, Srinadh Bhojanapalli, Anupam Gupta, and Chulhee Yun. 2024.
\newblock \href {https://arxiv.org/abs/2405.20671} {Position coupling: Improving length generalization of arithmetic transformers using task structure}.
\newblock \emph{Preprint}, arXiv:2405.20671.

\bibitem[{Cobbe et~al.(2021)Cobbe, Kosaraju, Bavarian, Chen, Jun, Kaiser, Plappert, Tworek, Hilton, Nakano, Hesse, and Schulman}]{cobbe2021trainingverifierssolvemath}
Karl Cobbe, Vineet Kosaraju, Mohammad Bavarian, Mark Chen, Heewoo Jun, Lukasz Kaiser, Matthias Plappert, Jerry Tworek, Jacob Hilton, Reiichiro Nakano, Christopher Hesse, and John Schulman. 2021.
\newblock \href {https://arxiv.org/abs/2110.14168} {Training verifiers to solve math word problems}.
\newblock \emph{Preprint}, arXiv:2110.14168.

\bibitem[{Cusanus(1440)}]{cusanus:1440}
Nicolas Cusanus. 1440.
\newblock \emph{De Docta Ignorantia}.
\newblock Google books reprinted in 2009.

\bibitem[{Davidson(1967)}]{davidson:1967}
Donald Davidson. 1967.
\newblock Truth and meaning.
\newblock \emph{Synthese}, 17:304--323.

\bibitem[{Deng et~al.(2026)Deng, Li, Xie, Chang, and Chen}]{deng:etal:2026}
Chunyuan Deng, Zhiqi Li, Roy Xie, Ruidi Chang, and Hanjie Chen. 2026.
\newblock \href {https://arxiv.org/abs/2410.15580} {Language models are symbolic learners in arithmetic}.
\newblock \emph{Preprint}, arXiv:2410.15580.

\bibitem[{Dummett(1993)}]{dummett:1993}
Michael Dummett. 1993.
\newblock \emph{The Seas of Language}.
\newblock Oxford University Press.

\bibitem[{Fel et~al.(2026)Fel, Wang, Lepori, Kowal, Lee, Balestriero, Joseph, Lubana, Konkle, Ba, and Wattenberg}]{fel:etal:2026}
Thomas Fel, Binxu Wang, Michael~A. Lepori, Matthew Kowal, Andrew Lee, Randall Balestriero, Sonia Joseph, Ekdeep~S. Lubana, Talia Konkle, Demba Ba, and Martin Wattenberg. 2026.
\newblock \href {https://arxiv.org/abs/2510.08638} {Into the rabbit hull: From task-relevant concepts in dino to minkowski geometry}.
\newblock \emph{Preprint}, arXiv:2510.08638.

\bibitem[{Feucht et~al.(2026)Feucht, Haklay, Bhalla, Wurgaft, Rager, Sarfati, Merullo, McGrath, Lewis, Lubana, Fel, and Geiger}]{feucht2026arithmeticwild}
Sheridan Feucht, Tal Haklay, Usha Bhalla, Daniel Wurgaft, Can Rager, Raphaël Sarfati, Jack Merullo, Thomas McGrath, Owen Lewis, Ekdeep~Singh Lubana, Thomas Fel, and Atticus Geiger. 2026.
\newblock \href {https://arxiv.org/abs/2605.01148} {Arithmetic in the wild: Llama uses base-10 addition to reason about cyclic concepts}.
\newblock \emph{Preprint}, arXiv:2605.01148.

\bibitem[{Fibonacci(1202)}]{fibonacci:1202}
Leonardo Fibonacci. 1202.
\newblock \emph{Liber Abaci}.
\newblock republished and translated by Jordi Deulofeu Piquet et Stephen Sanchez (Trad.), Le créateur de la suite mathématique de la beauté : Fibonacci, Barcelone, RBA Coleccionables, 2018.

\bibitem[{Golkar et~al.(2024)Golkar, Pettee, Eickenberg, Bietti, Cranmer, Krawezik, Lanusse, McCabe, Ohana, Parker, Blancard, Tesileanu, Cho, and Ho}]{golkar2024xval}
Siavash Golkar, Mariel Pettee, Michael Eickenberg, Alberto Bietti, Miles Cranmer, Geraud Krawezik, Francois Lanusse, Michael McCabe, Ruben Ohana, Liam Parker, Bruno Régaldo-Saint Blancard, Tiberiu Tesileanu, Kyunghyun Cho, and Shirley Ho. 2024.
\newblock \href {https://arxiv.org/abs/2310.02989} {xval: A continuous numerical tokenization for scientific language models}.
\newblock \emph{Preprint}, arXiv:2310.02989.

\bibitem[{Gouvert et~al.(2025)Gouvert, Hunter, Louradour, Cerisara, Dufraisse, Sy, Rivière, Lorré, and community}]{openllm2025lucie}
Olivier Gouvert, Julie Hunter, Jérôme Louradour, Christophe Cerisara, Evan Dufraisse, Yaya Sy, Laura Rivière, Jean-Pierre Lorré, and OpenLLM-France community. 2025.
\newblock \href {https://arxiv.org/abs/2503.12294} {The lucie-7b llm and the lucie training dataset: Open resources for multilingual language generation}.
\newblock \emph{Preprint}, arXiv:2503.12294.

\bibitem[{Hou et~al.(2024)Hou, Brandfonbrener, Kakade, Jelassi, and Malach}]{hou:etal:2024}
Kaiying Hou, David Brandfonbrener, Sham Kakade, Samy Jelassi, and Eran Malach. 2024.
\newblock \href {https://arxiv.org/abs/2407.03310} {Universal length generalization with turing programs}.
\newblock \emph{Preprint, arXiv:2407.03310}.

\bibitem[{Hupkes et~al.(2023)Hupkes, Giulianelli, Dankers, Artetxe, Elazar, Pimentel, Christodoulopoulos, Lasri, Saphra, Sinclair, Ulmer, Schottmann, Batsuren, Sun, Sinha, Khalatbari, Ryskina, Frieske, Cotterell, and Jin}]{hupkes:etal:2023}
Dieuwke Hupkes, Mario Giulianelli, Verna Dankers, Mikel Artetxe, Yanai Elazar, Tiago Pimentel, Christos Christodoulopoulos, Karim Lasri, Naomi Saphra, Arabella Sinclair, Dennis Ulmer, Florian Schottmann, Khuyagbaatar Batsuren, Kaiser Sun, Koustuv Sinha, Leila Khalatbari, Maria Ryskina, Rita Frieske, Ryan Cotterell, and Zhijing Jin. 2023.
\newblock A taxonomy and review of generalization research in {NLP}.
\newblock \emph{Nature Machine Intelligence}, 5:1161--1174.

\bibitem[{Karkada et~al.(2026)Karkada, Korchinski, Nava, Wyart, and Bahri}]{karkada:etal:2026}
Dhruva Karkada, Daniel~J Korchinski, Andres Nava, Matthieu Wyart, and Yasaman Bahri. 2026.
\newblock Symmetry in language statistics shapes the geometry of model representations.
\newblock \emph{arXiv preprint arXiv:2602.15029}.

\bibitem[{Keysers et~al.(2020)Keysers, Sch{\"a}rli, Scales, Buisman, Furrer, Kashubin, Momchev, Sinopalnikov, Stafiniak, Tihon, Tsarkov, Wang, van Zee, and Bousquet}]{keysers:etal:2020}
Daniel Keysers, Nathanael Sch{\"a}rli, Nathan Scales, Hylke Buisman, Daniel Furrer, Sergii Kashubin, Nikola Momchev, Danila Sinopalnikov, Lukasz Stafiniak, Tibor Tihon, Dmitry Tsarkov, Xiao Wang, Marc van Zee, and Olivier Bousquet. 2020.
\newblock Measuring compositional generalization: A comprehensive method on realistic data.
\newblock In \emph{International Conference on Learning Representations (ICLR)}.

\bibitem[{Kim and Linzen(2020)}]{kim:linzen:2020}
Najoung Kim and Tal Linzen. 2020.
\newblock {COGS}: A compositional generalization challenge based on semantic interpretation.
\newblock In \emph{Proceedings of the 2020 Conference on Empirical Methods in Natural Language Processing (EMNLP)}, pages 9087--9105.

\bibitem[{Kripke(1982)}]{kripke:1982}
Saul~A Kripke. 1982.
\newblock \emph{Wittgenstein on Rules and Private Language: An Elementary Exposition}.
\newblock Harvard University Press.

\bibitem[{Lake and Baroni(2018)}]{lake:baroni:2018}
Brenden~M. Lake and Marco Baroni. 2018.
\newblock Generalization without systematicity: On the compositional skills of sequence-to-sequence recurrent networks.
\newblock In \emph{Proceedings of the 35th International Conference on Machine Learning (ICML)}.

\bibitem[{Lampinen(2024)}]{lampinen:2024}
Andrew~K. Lampinen. 2024.
\newblock \href {https://doi.org/10.1162/coli_a_00525} {Can language models handle recursively nested grammatical structures? a case study on comparing models and humans}.
\newblock \emph{Computational Linguistics}, 50(4).

\bibitem[{{Llama Team}(2024)}]{llama3herd}
{Llama Team}. 2024.
\newblock \href {https://doi.org/10.48550/ARXIV.2407.21783} {The llama 3 herd of models}.
\newblock \emph{CoRR}, abs/2407.21783.

\bibitem[{Lonergan(1958)}]{lonergan:1958}
Bernard Lonergan. 1958.
\newblock \emph{Insight}.
\newblock Harper \& Row Publishers.

\bibitem[{McCoy et~al.(2024)McCoy, Yao, Friedman, Hardy, and Griffiths}]{mccoy2024embers}
R~Thomas McCoy, Shunyu Yao, Dan Friedman, Mathew~D Hardy, and Thomas~L Griffiths. 2024.
\newblock Embers of autoregression show how large language models are shaped by the problem they are trained to solve.
\newblock \emph{Proceedings of the National Academy of Sciences}, 121(41):e2322420121.

\bibitem[{McGinn(1984)}]{mcginn:1984}
Colin McGinn. 1984.
\newblock \emph{Wittgenstein on Meaning}.
\newblock Basil Blackwell Publishers.

\bibitem[{McLeish et~al.(2024)McLeish, Bansal, Stein, Jain, Kirchenbauer, Bartoldson, Kailkhura, Bhatele, Geiping, Schwarzschild, and Goldstein}]{mcleish2024abacus}
Sean McLeish, Arpit Bansal, Alex Stein, Neel Jain, John Kirchenbauer, Brian~R. Bartoldson, Bhavya Kailkhura, Abhinav Bhatele, Jonas Geiping, Avi Schwarzschild, and Tom Goldstein. 2024.
\newblock \href {https://arxiv.org/abs/2405.17399} {Transformers can do arithmetic with the right embeddings}.
\newblock \emph{Preprint}, arXiv:2405.17399.

\bibitem[{Miao et~al.(2020)Miao, Liang, and Su}]{miao-etal-2020-diverse}
Shen-yun Miao, Chao-Chun Liang, and Keh-Yih Su. 2020.
\newblock \href {https://doi.org/10.18653/v1/2020.acl-main.92} {A diverse corpus for evaluating and developing {E}nglish math word problem solvers}.
\newblock In \emph{Proceedings of the 58th Annual Meeting of the Association for Computational Linguistics}, pages 975--984, Online. Association for Computational Linguistics.

\bibitem[{Muffo et~al.(2022)Muffo, Cocco, and Bertino}]{muffo2022}
Matteo Muffo, Aldo Cocco, and Enrico Bertino. 2022.
\newblock \href {https://aclanthology.org/2022.lrec-1.30/} {Evaluating transformer language models on arithmetic operations using number decomposition}.
\newblock In \emph{Proceedings of the Thirteenth Language Resources and Evaluation Conference}, pages 291--297, Marseille, France. European Language Resources Association.

\bibitem[{Naim et~al.(2026)Naim, Bhar, Bolte, and Asher}]{naim:etal:2026}
Omar Naim, Swarnadeep Bhar, Jérôme Bolte, and Nicholas Asher. 2026.
\newblock \href {https://arxiv.org/abs/2508.14685} {Ssa: Improving performance with a better scoring function}.
\newblock \emph{Preprint}, arXiv:2508.14685.

\bibitem[{Naim et~al.(2025{\natexlab{a}})Naim, Bolte, and Asher}]{naim:etal:2025}
Omar Naim, J{\'e}r{\^o}me Bolte, and Nicholas Asher. 2025{\natexlab{a}}.
\newblock Analyzing limits for in-context learning.
\newblock \emph{arXiv preprint arXiv:2502.03503}.

\bibitem[{Naim et~al.(2025{\natexlab{b}})Naim, Fouilh{\'e}, and Asher}]{naim2025re}
Omar Naim, Guilhem Fouilh{\'e}, and Nicholas Asher. 2025{\natexlab{b}}.
\newblock Re-examining learning linear functions in context.
\newblock In \emph{German Conference on Artificial Intelligence (K{\"u}nstliche Intelligenz)}, pages 104--117. Springer.

\bibitem[{Nikankin et~al.(2025)Nikankin, Reusch, Mueller, and Belinkov}]{nikankin:etal:2025}
Yaniv Nikankin, Anja Reusch, Aaron Mueller, and Yonatan Belinkov. 2025.
\newblock \href {https://openreview.net/forum?id=O9YTt26r2P} {Arithmetic without algorithms: Language models solve math with a bag of heuristics}.
\newblock In \emph{The Thirteenth International Conference on Learning Representations (ICLR)}.

\bibitem[{Nye et~al.(2021)Nye, Andreassen, Gur-Ari, Michalewski, Austin, Bieber, Dohan, Lewkowycz, Bosma, Luan, Sutton, and Odena}]{nye2021scratchpads}
Maxwell Nye, Anders~Johan Andreassen, Guy Gur-Ari, Henryk Michalewski, Jacob Austin, David Bieber, David Dohan, Aitor Lewkowycz, Maarten Bosma, David Luan, Charles Sutton, and Augustus Odena. 2021.
\newblock \href {https://arxiv.org/abs/2112.00114} {Show your work: Scratchpads for intermediate computation with language models}.
\newblock \emph{Preprint}, arXiv:2112.00114.

\bibitem[{Olsson et~al.(2022)Olsson, Elhage, Nanda, Joseph, DasSarma, Henighan, Mann, Askell, Bai, Chen et~al.}]{olsson:etal:2022}
Catherine Olsson, Nelson Elhage, Neel Nanda, Nicholas Joseph, Nova DasSarma, Tom Henighan, Ben Mann, Amanda Askell, Yuntao Bai, Anna Chen, and 1 others. 2022.
\newblock In-context learning and induction heads.
\newblock \emph{arXiv preprint arXiv:2209.11895}.

\bibitem[{Osherson and Weinstein(1986)}]{osherson:weinstein:1986}
D.~Osherson and S.~Weinstein. 1986.
\newblock Identification in the limit of first order structures.
\newblock \emph{Journal of Philosophical Logic}, 15:55–81.

\bibitem[{P{\'e}rez et~al.(2021)P{\'e}rez, Barcel{\'o}, and Marinkovic}]{perez:etal:2021}
Jorge P{\'e}rez, Pablo Barcel{\'o}, and Javier Marinkovic. 2021.
\newblock Attention is turing-complete.
\newblock \emph{Journal of Machine Learning Research}, 22(75):1--35.

\bibitem[{Pinker and Prince(1988)}]{pinker:prince:1988}
Steven Pinker and Alan Prince. 1988.
\newblock On language and connectionism: Analysis of a parallel distributed processing model of language acquisition.
\newblock \emph{Cognition}, 28(1-2):73--193.

\bibitem[{Pouget et~al.(2000)Pouget, Dayan, and Zemel}]{pouget:etal:2000}
Alexandre Pouget, Peter Dayan, and Richard Zemel. 2000.
\newblock Information processing with population codes.
\newblock \emph{Nature Reviews Neuroscience}, 1(2):125--132.

\bibitem[{Qian et~al.(2022)Qian, Wang, Li, Li, and Yan}]{qian2023}
Jing Qian, Hong Wang, Zekun Li, Shiyang Li, and Xifeng Yan. 2022.
\newblock \href {https://arxiv.org/abs/2208.05051} {Limitations of language models in arithmetic and symbolic induction}.
\newblock \emph{Preprint}, arXiv:2208.05051.

\bibitem[{Qiu et~al.(2022)Qiu, Shaw, Pasupat, Nowak, Linzen, Sha, and Toutanova}]{qiu:etal:2022}
Linlu Qiu, Peter Shaw, Panupong Pasupat, Pawel Nowak, Tal Linzen, Fei Sha, and Kristina Toutanova. 2022.
\newblock Improving compositional generalization with latent structure and data augmentation.
\newblock In \emph{Proceedings of the 2022 conference of the north american chapter of the association for computational linguistics: Human language technologies}, pages 4341--4362.

\bibitem[{Quine(1960)}]{quine:1960}
Willard Van~Orman Quine. 1960.
\newblock \emph{Word and object}.
\newblock Harvard University Press.

\bibitem[{{Qwen Team}(2025)}]{qwen3technicalreport}
{Qwen Team}. 2025.
\newblock \href {https://arxiv.org/abs/2505.09388} {Qwen3 technical report}.
\newblock \emph{Preprint}, arXiv:2505.09388.

\bibitem[{Radford et~al.(2019)Radford, Wu, Child, Luan, Amodei, Sutskever et~al.}]{radford2019language}
Alec Radford, Jeffrey Wu, Rewon Child, David Luan, Dario Amodei, Ilya Sutskever, and 1 others. 2019.
\newblock Language models are unsupervised multitask learners.
\newblock \emph{OpenAI blog}, 1(8):9.

\bibitem[{Rumelhart and McClelland(1986)}]{rumelhart:mcclelland:1986}
D.~E. Rumelhart and J.~L. McClelland. 1986.
\newblock \href {https://doi.org/10.7551/mitpress/5237.003.0008} {On learning the past tenses of english verbs}.
\newblock In \emph{Parallel Distributed Processing, Volume 2: Explorations in the Microstructure of Cognition: Psychological and Biological Models}. The MIT Press.

\bibitem[{Shoenfield(1959)}]{schoenfeld:1959}
J.~R. Shoenfield. 1959.
\newblock On degrees of unsolvability.
\newblock \emph{Annals of Mathematics}, 69:644–653.

\bibitem[{Sinclair et~al.(2022)Sinclair, Jumelet, Zuidema, and Fern{\'a}ndez}]{sinclair:etal:2022}
Arabella Sinclair, Jaap Jumelet, Willem Zuidema, and Raquel Fern{\'a}ndez. 2022.
\newblock \href {https://doi.org/10.1162/tacl_a_00504} {Structural persistence in language models: Priming as a window into abstract language representations}.
\newblock \emph{Transactions of the Association for Computational Linguistics}, 10:1031--1050.

\bibitem[{Tong et~al.(2026)Tong, Ye, Borovykh, and Shokri}]{nr_shortest}
Yao Tong, Jiayuan Ye, Anastasia Borovykh, and Reza Shokri. 2026.
\newblock \href {https://arxiv.org/abs/2604.15306} {Generalization in llm problem solving: The case of the shortest path}.
\newblock \emph{Preprint}, arXiv:2604.15306.

\bibitem[{Wittgenstein(1953)}]{wittgenstein:1953}
Ludwig Wittgenstein. 1953.
\newblock \emph{Philosophical investigations}.
\newblock Blackwells.

\bibitem[{Xie et~al.(2021)Xie, Raghunathan, Liang, and Ma}]{xie:etal:2021}
Sang~Michael Xie, Aditi Raghunathan, Percy Liang, and Tengyu Ma. 2021.
\newblock An explanation of in-context learning as implicit bayesian inference.
\newblock \emph{arXiv preprint arXiv:2111.02080}.

\bibitem[{Yan et~al.(2025)Yan, Lu, Xu, and Lan}]{yan2025}
Yang Yan, Yu~Lu, Renjun Xu, and Zhenzhong Lan. 2025.
\newblock \href {https://doi.org/10.18653/v1/2025.emnlp-main.681} {Do large language models truly grasp addition? a rule-focused diagnostic using two-integer arithmetic}.
\newblock In \emph{Proceedings of the 2025 Conference on Empirical Methods in Natural Language Processing}, pages 13467--13483, Suzhou, China. Association for Computational Linguistics.

\bibitem[{Zhang et~al.(2024)Zhang, Tigges, Zhang, Biderman, Raginsky, and Ringer}]{nr_recursion}
Dylan Zhang, Curt Tigges, Zory Zhang, Stella Biderman, Maxim Raginsky, and Talia Ringer. 2024.
\newblock \href {https://arxiv.org/abs/2401.12947} {Transformer-based models are not yet perfect at learning to emulate structural recursion}.
\newblock \emph{Preprint}, arXiv:2401.12947.

\bibitem[{Zhang et~al.(2022)Zhang, Roller, Goyal, Artetxe, Chen, Chen, Dewan, Diab, Li, Lin, Mihaylov, Ott, Shleifer, Shuster, Simig, Koura, Sridhar, Wang, and Zettlemoyer}]{zhang2022opt}
Susan Zhang, Stephen Roller, Naman Goyal, Mikel Artetxe, Moya Chen, Shuohui Chen, Christopher Dewan, Mona Diab, Xian Li, Xi~Victoria Lin, Todor Mihaylov, Myle Ott, Sam Shleifer, Kurt Shuster, Daniel Simig, Punit~Singh Koura, Anjali Sridhar, Tianlu Wang, and Luke Zettlemoyer. 2022.
\newblock \href {https://arxiv.org/abs/2205.01068} {Opt: Open pre-trained transformer language models}.
\newblock \emph{Preprint}, arXiv:2205.01068.

\bibitem[{Zhou et~al.(2024)Zhou, Fu, Sharan, and Jia}]{zhou2024fourier}
Tianyi Zhou, Deqing Fu, Vatsal Sharan, and Robin Jia. 2024.
\newblock \href {https://arxiv.org/abs/2406.03445} {Pre-trained large language models use fourier features to compute addition}.
\newblock \emph{Preprint}, arXiv:2406.03445.

\bibitem[{Zhou et~al.(2026)Zhou, Fu, Soltanolkotabi, Jia, and Sharan}]{zhou2026fone}
Tianyi Zhou, Deqing Fu, Mahdi Soltanolkotabi, Robin Jia, and Vatsal Sharan. 2026.
\newblock \href {https://arxiv.org/abs/2502.09741} {Fone: Precise single-token number embeddings via fourier features}.
\newblock \emph{Preprint}, arXiv:2502.09741.

\end{thebibliography}
\clearpage

\appendix
\raggedbottom

\section{Learning in the limit}
\label{sec:limits}
Here we investigate {\em learning in the limit} or {\em limit computability}, which applies to infinitary concepts.  
\begin{definition} \cite{schoenfeld:1959} A set $S \subseteq \mathbb{N}$ is limit computable if for $x, s \in \mathbb{N}$ its characteristic function 
$$\phi_S(x) =  lim_{s \rightarrow \omega}g(x,s)$$ for some total computable function $g(x,s)$.
\end{definition}

Since the Kripkenstein arguments attack the possibility of justifiably inferring a universal rule from finite instances, let us consider universal quantification and learnability in the limit. Universal quantification is a function over pairs of properties or sets into $\{0, 1\}$ such that 
$$f(P,Q) = 1 \  \text{ iff } \  P \subseteq Q$$
We consider an arithmetic version of universal quantification in which we set $P = \mathbb{N}$.  We can generalize this argument to arbitrary countable models $M$, by ``linearizing'' the domain (assigning each element a natural number) and treat $s$ a sequence of objects as characterizing a finite submodel of $M$.  $M$ is then characterized by an infinite sequence.  Quantified statements then correspond to countable unions and intersections of such sequences, creating the Borel Hierarchy.

There is a computable function g(s,x) where $x$ ranges over elements of $s$ and $s$ is a finite initial segment of $\omega$ and
$$ g(x,s) = 1 \mbox { iff } \forall x \in s (x \in P \rightarrow x \in Q) $$ 
  Over $\mathbb{N}$, for each $s < \omega$, $g(s,x)$ is computable.  However, $\{g(s,x)\}_{s \in \omega}$ doesn't necessarily converge to $0$ or $1$ on $\omega$.  

As \cite{osherson:weinstein:1986} remark however, a universal statement is computable in the limit  with "one hypothesis change" ({\em co-computably enumerable} in their terminology).  That is, we can learn a universal claim $
\phi$ if we allow one revision to our hypothesis about $\phi$: we start out with the assumption that $\phi$ {\em is true}, and if we encounter a falsifying instance then we conclude that $\phi$ is false.  Conversely, we can learn in the limit an existential statement $\psi$ over $\mathbb{N}$ with one hypothesis change, if we start out with the assumption that $\psi$ is false.  
\begin{proposition}
$\phi$ is computable in the limit iff it is computably enumerable and co-computably enumerable.
\end{proposition}
\noindent
An adaptation of the theorem that a set is computable iff it is recursively enumerable and co recursively enumerable proves this proposition.

{\bf Evaluation.}  Learnability in the limit arguments depend on the assumption that terms have {\em a constant meaning as we evaluate a universal or existential formula in ever larger finite models. }  But this is precisely what the skeptical attack on meanings disputes.  

The Kripkenstein argument gives us a constraint on probability distributions. $P$ obeys a {\em Kripkean justification pattern}, if $P(x|\sigma) \approx P(y|\sigma)$ for OOD $x,y$ and $\sigma$ in ID.
\begin{proposition}
$\Pi_1, \Sigma_1$ sets are not learnable in the limit with 1 hypothesis change, given objective functions with Kripkean justification patterns. \cite{asher:etal:2023}
\end{proposition}

\section{Six arithmetical experiments}\label{app:arithmetic}

\subsection{Experimental Setup}

We evaluate addition under digit length shifts. For a model with training range $1,\ldots n$, in distribution examples use operands with at most $n$ digits. Near OOD examples extend the operand length slightly beyond the training range and far OOD examples extend it further. We report exact match accuracy on the full generated answer, since digit level metrics can obscure correlated errors across output positions.

Our experiments cover several pretrained causal language models: GPT-2 small \citep{radford2019language}, OPT-125M \citep{zhang2022opt}, SmolLM-135M \citep{allal2024SmolLM}, Qwen3-0.6B \citep{qwen3technicalreport}, Llama-3.2-1B, Llama-3.1-70B \citep{llama3herd}, and Lucie-7B \citep{openllm2025lucie}. 
 We vary the form of structure added to the model: numerical representation (\S\ref{sec:interfaces}), positional scaffolding (\S\ref{sec:digit-shift}), an explicit rule supplied at inference time (\S\ref{sec:prompted-rule}), and procedural supervision on intermediate states (\S\ref{sec:structure}).


\subsection{Structured numerical interfaces}
\label{sec:interfaces}
Prior work has shown that specialized numerical interfaces, including continuous number tokenization, digit-position embeddings, and Fourier number embeddings, can improve numerical reasoning over ordinary text tokenization \citep{golkar2024xval,mcleish2024abacus,zhou2026fone}. Motivated by this, we compare ordinary text fine-tuning with two structured numerical interfaces.

The first is a \emph{column-state decoder}.\footnote{Related to position-coupling \citep{cho2024positioncoupling}, but with explicit column-indexed digit states instead of aligned digit positional encodings.} Each number is replaced in the prompt with a special numerical token, while its integer value is passed through a structured numerical encoder, producing one embedding for the whole number. We add a dedicated number decoder for these embeddings.

The second is an \emph{auxiliary supervised text method}. Here the model keeps the ordinary language-model output head, and answers are still generated as text tokens. Unlike the column-state decoder, the numerical pathway does not replace the text pathway: numerical embeddings are added to ordinary token embeddings at number positions, so the transformer receives both the pretrained text representation and explicit numerical structure.

The distinction is deliberate. For smaller models we replace the weak native numerical pathway, while for stronger pretrained models we use a lighter auxiliary interface motivated by FoNE-style features \citep{zhou2026fone} that preserves the BPE/text pathway. Table~\ref{tab:interface-example} shows the three interfaces on a single problem; implementation details are in Appendix~\ref{app:numeric-implementation}.

\begin{table}[h]
\centering
\small
\setlength{\tabcolsep}{4pt}
\resizebox{\linewidth}{!}{
\begin{tabular}{lll}
\toprule
& \textbf{Model input} & \textbf{Answer produced by} \\
\midrule
Plain & \texttt{372 + 158 =} & LM head, text tokens \texttt{530} \\[2pt]
Col   & \texttt{<num> + <num> =} & number decoder, LSB-first \\
      & \quad values $372,158$ via encoder & \quad $0,3,5 \rightarrow 530$ \\
      & \quad columns $(2,7,3),(8,5,1)$ & \\[2pt]
Aux   & \texttt{372 + 158 =} & LM head, text tokens \texttt{530} \\
      & \quad $e'_t = e_t + g\cdot\mathrm{num}(x)$ & \\
\bottomrule
\end{tabular}}
\caption{The same problem under the three interfaces. Col changes the numeric interface, replacing numbers in the text stream and decoding the answer outside the LM head. Aux leaves that interface intact and changes only what the hidden states are encouraged to represent.}
\label{tab:interface-example}
\end{table}

\hidden{
Prior work has shown that specialized numerical interfaces, including continuous number tokenization, digit-position embeddings, and Fourier number embeddings, can substantially improve numerical reasoning compared with ordinary text tokenization \citep{golkar2024xval,mcleish2024abacus,zhou2026fone}. Motivated by this, we compare ordinary text fine-tuning with two structured numerical interfaces. The first is a \emph{column-state decoder}. \footnote{Related to position-coupling \citep{cho2024positioncoupling}, but with explicit column-indexed digit states instead of aligned digit positional encodings.} In this setting, 
each number is replaced in the prompt with a special numerical token, while its integer value is passed through a structured numerical encoder, producing one embedding for the whole number. 
We add a dedicated number decoder for these new embeddings (details are in Appendix \ref{app:numeric-implementation}).

The second interface is an \emph{auxiliary supervised text method}. Here the model keeps the ordinary language-model output head. Answers are still generated as text tokens. Unlike the column-state decoder, the numerical pathway does not replace the text pathway. Instead, numerical embeddings are added to the ordinary token embeddings at number positions. So the transformer receives both the pretrained text representation and explicit numerical  representation (details in Appendix \ref{app:numeric-implementation}). 

The distinction is deliberate. For smaller models, we test whether replacing the weak native numerical pathway with an explicit column-state decoder improves rule extension. For stronger pretrained models, we use a lighter auxiliary interface, motivated by FoNE-style numerical features \citep{zhou2026fone}, that preserves the existing BPE/text pathway while adding explicit numerical structure. Additional details about the data construction, numerical encoder, output decoder, losses and training procedure are given in the Appendix~\ref{app:numeric-implementation}.
}
\subsection{Addition under digit-length shift}
\label{sec:digit-shift}
\begin{table}[h]
\centering
\setlength{\tabcolsep}{4pt}
\resizebox{\linewidth}{!}{
\begin{tabular}{lccccc}
\toprule
 Backbone & $n$ & Method & ID & Near-OOD & Far-OOD \\
\midrule
   GPT-2 small & 6 & Plain & 30.18\% & 0.00\% & 0.00\% \\
   GPT-2 small & 6 & Col  & 99.50\% & 58.34\% & 1.36\% \\
   OPT-125M & 6 & Plain & 36.42\% & 0.00\% & 0.00\% \\
   OPT-125M & 6  & Col  & 100.00\% & 58.96\% & 1.47\% \\
   SmolLM-135M & 6  & Plain & 94.67\% & 15.85\% & 0.00\% \\
   SmolLM-135M & 6 & Col  & 99.30\% & 59.27\% & 1.47\% \\
   Llama-3.2-1B & 15 & Plain & 94.44\% & 0.10\% & 0.00\% \\
   Llama-3.2-1B & 15  & Aux  & 97.69\% & 26.65\% & 0.10\% \\
   Qwen3-0.6B & 20 & Plain & 76.12\% & 0.00\% & 0.00\% \\
   Qwen3-0.6B & 20 & Aux & 98.50\% & 73.40\% & 10.35\% \\
   Lucie-7B & 30 & Plain & 98.60\% & 13.60\% & 2.29\% \\
   Lucie-7B & 30 & Aux & 99.11\% & 33.55\% & 8.18\% \\
  \bottomrule
\end{tabular}}
\caption{Addition under digit-length shift. $n$ is the maximum operand digit length seen in training; ID uses operands of at most $n$ digits, Near-OOD uses $n{+}1,n{+}2$, and Far-OOD uses $n{+}3,n{+}4$.}
\label{tab:id-ood-tab}
\end{table}

Table~\ref{tab:id-ood-tab} shows a consistent pattern. Plain text models can achieve high in-distribution accuracy, but often fail sharply when answer length moves beyond the training regime. Structured numerical interfaces substantially improve in-distribution and near-OOD accuracy, showing that the models can exploit useful arithmetic structure with a more favorable representation. 

The improvement, however, is not equivalent to acquiring the addition rule. For the column-state and auxiliary interfaces in Table~\ref{tab:id-ood-tab}, far-OOD addition remains largely unsolved for our tested interfaces. Our point is not that numerical structure is useless, but that better numerical representations and supervised readout, while improving local generalization by themselves, do not specify how the rule should extend.

The interfaces above change how numbers are represented. A different kind of intervention changes how positions are indexed. We therefore tested an Abacus-style \citep{mcleish2024abacus} positional scaffold on Qwen3-0.6B, which supplies an explicit coordinate for digit-column alignment.

\begin{table}[h]
\centering
\setlength{\tabcolsep}{4pt}
\resizebox{\linewidth}{!}{
\begin{tabular}{lccccc}
\toprule
Model & Method & $n$ & ID & Near-OOD & Far-OOD \\
\midrule
Qwen3-0.6B & Plain & 20 & 76.12\% & 0.00\% & 0.00\% \\
Qwen3-0.6B & Abacus & 20 & 91.80\% & 72.85\% & 34.20\% \\
\bottomrule
\end{tabular}}
\caption{Qwen3-0.6B addition results comparing plain with an abacus-style positional scaffold}
\label{tab:qwen-abacus}
\end{table}

We take Table~\ref{tab:qwen-abacus} as a positive result. Far-OOD addition is not beyond transformer models, and different forms of structure extend the generalization range to markedly different degrees. This supports rather than weakens the rule-following diagnosis. The Abacus scaffold does not encode the addition algorithm: it supplies no carry computation and no column-update procedure. What it supplies is an inductive bias constraining how the finite training examples are to be extended, and that bias is doing most of the work. At the same time, a substantial gap remains 
even in the strongest condition. A coordinate system that determines how digits align is not the same as a procedure that executes reliably as the domain grows.

\subsection{Good representation $\neq$ rule following}

Recent mechanistic work suggests that language models often represent numbers through periodic or Fourier-like features, where a number is encoded by its residues modulo periods such as 2, 5, 10, or 100 \citep{zhou2024fourier, feucht2026arithmeticwild}. Relatedly, \citet{nikankin:etal:2025} find that language models solve arithmetic through a bag of heuristics implemented by distributed neuron sets rather than a general algorithm, and \citet{deng:etal:2026} report converging evidence on the robustness of model arithmetic. These accounts are consistent with our diagnosis; our question is what follows for how competence is assessed. These periods are natural for base-10 arithmetic: for example, $n \bmod 10$ gives the final digit of $n$, and $(a+b)\bmod 10$ gives the final digit of the sum. We used a linear probe to see whether the hidden state at the answer position contains residues of the operands and of the sum. {Table~\ref{tab:residue-probe} reports the modulo-10 probe}; Appendix~\ref{app:residue-probes} reports the corresponding probes for other periods.

\begin{table}[h]
\centering
\resizebox{\linewidth}{!}{
\begin{tabular}{lcccc}
\toprule
    Model & Method & $a\bmod 10$ & $b\bmod 10$ & $(a+b)\bmod 10$ \\
    \midrule
    Llama-3.2-1B & plain & 87.0\% & 98.1\% & 12.7\% \\
    Llama-3.2-1B & aux   & 99.6\% & 99.8\% & 84.8\%  \\
\bottomrule
\end{tabular}}
\caption{Linear-probe accuracy for base-10 residue information at the answer-context hidden state.}
\label{tab:residue-probe}
\end{table}

In a plain arithmetic-finetuned Llama-3.2-1B model, operand residues such as $a\bmod 10$ and $b\bmod 10$ are decodable, but the sum residue $(a+b)\bmod 10$ is nearly absent. With our numerical-auxiliary training, $(a+b)\bmod 10$ becomes highly decodable from the answer-context hidden state. However, this representational improvement does not translate into stable length generalization: exact-match accuracy still collapses with addition of numbers with more digits. It is not that the model just lacks numerical information internally. Rather, finite training can induce local arithmetic geometry without yielding a rule-like procedure for extending addition beyond the training range. 

We state the conclusion narrowly. Digit-length shift tests two things at once: whether a model can represent numerals of increasing length, and whether it can apply the addition procedure to them. Our probes do not fully separate the two. They show that better internal access to numerical information is not sufficient for stable extension, but not that numeral representation itself has been solved.

\subsection{Giving the rule doesn't yield rule following}
\label{sec:prompted-rule}

If improved numerical representations do not by themselves yield rule-like extensions, perhaps model failure is due to the relevant rule not being explicitly available to the model. 

We tested a prompt-only setting in which the model is given the semi-decision procedure for universal statements from the learning-in-the-limit discussion in Appendix~\ref{sec:limits}: assume the statement is true, scan the sequence from left to right, and switch to false once a counterexample is found. We compare a task-only prompt with a rule-guided prompt that states this scanning algorithm and evaluate Llama-3.1-70B-Instruct on sequences of increasing length for the question: \emph{Are all numbers in the sequence positive?} Details are given in Appendix~\ref{app:algo}.

The rule-guided prompt improves performance on short and moderate sequence lengths but does not eliminate the length cliff: exact-match accuracy still collapses as sequence length increases.  This matters for our argument because it separates possession of a rule description from stable rule application. A model can be told the relevant rule and still fail to execute it reliably under domain extension. The issue is therefore not merely whether the rule is expressible in the prompt, but whether the system has a stable procedure for applying it as the domain grows.


\hidden{

{\color{red} Niyar maybe modify starting from this paragraph and merge it with your paragraph We evaluate addition under digit-length shifts. Models are trained or fine-tuned on addition problems within a bounded digit range and tested on three regimes: in-distribution examples, whose digit lengths match the training range; near-OOD examples, whose digit lengths extend the range slightly; and far-OOD examples, which extend it further. We compare plain text fine-tuning with two structured variants. The column-state method replaces input numbers with special numericaltokens, passes their values through a structured numericalencoder, and uses a digit decoder conditioned on the transformer state to emit the answer least-significant digit first. The auxiliary supervised method keeps generation in text, but adds intermediate supervision designed to make arithmetic structure more explicit during training. Details of data construction, training, decoding, and model configurations are given in Appendix~XXXX.

Table~\ref{tab:id-ood-tab} shows the resulting pattern. Plain text models generally fail to generalize beyond the training range. Structured methods substantially improve in-distribution and near-OOD accuracy, showing that the models can exploit useful arithmetic structure when the representation is made more favorable. Nevertheless, far-OOD addition remains largely unsolved. For example, column-state models trained on one- to six-digit additions reach nearly perfect in-distribution accuracy, but drop to roughly one percent accuracy on far-OOD inputs. Larger models with auxiliary supervision show the same qualitative pattern: performance improves near the training range, but exact-match accuracy still collapses under larger digit-length shifts.
}

Do current language models have objective functions with Kripkean justification patterns?  There is some evidence in the literature for a "yes" answer.  \cite{naim:etal:2025} shows that small GPT2 style models trained from scratch on linear and polynomial regression tasks have excellent performance when test inputs match in distribution training data characteristics, but they fail in rather dramatic fashion once the test distribution is shifted significantly from in-distribution.  We see similar behavior for small and midsized models (Llama 3 70b) on quantification \cite{naim:etal:2026} (see Figures \ref{hmapOR} and \ref{hmapAND}).  \textcolor{red}{add these citations} \textcolor{violet}{In our arithmetic experiments, we evaluate this behavior across several pretrained causal language models: GPT-2 small \citep{}, OPT-125M \citep{}, SmolLM-135M \citep{}, Qwen3-0.6B \citep{}, Llama-3.2-1B and Llama-3.1-8B \citep{}, and Lucie-7B \citep{}.} Data about addition also reveal a "cliff-like" failure mode on OOD inputs.  Small models trained on arithmetic tasks with specialized encodings perform with 98\% accuracy or higher on ID data with 6 digit numbers but get 0 \% accuracy or close to it with 7 digit numbers.  As shown in Table~\ref{tab:id-ood-tab}, larger language models in the 0.6B--7B parameter range can achieve near-perfect in-distribution addition accuracy for operands up to 15--30 digits, yet their performance drops sharply once the required answer length exceeds the training range.

\textcolor{violet}{The column-state rows of Table \ref{tab:id-ood-tab} should not be read as results from a frozen external calculator attached to a language model. The pretrained language-model backbone, numericalencoder-adapter, and digit decoder are fine-tuned jointly. Input numbers are replaced by a special numericaltoken, while their values are passed through a structured numericalencoder and injected into the transformer through an adapter. At inference time, numericalanswers are not produced by the standard language-model head. Instead, the final context state before the answer conditions a column-state decoder, which emits the result least-significant digit first. Frozen-backbone ablations perform substantially worse, suggesting that the useful numerical structure must be integrated into the model’s internal representations rather than attached as a post-hoc module. The important point for our argument is that even this integrated numerical pathway improves only near-OOD generalization; far-OOD addition remains essentially unsolved. }
}
\subsection{Providing structure in training}
\label{sec:structure}

The prompt experiment gives the model the rule at inference time. A stronger intervention is to train the model on the intermediate states of the computation itself. This is close to the Turing--program style scratchpad supervision, where an algorithm is decomposed into local state updates that mimic a Turing--machine computation \cite{hou:etal:2024}.  For universal quantification, the intermediate state is the running truth value of the universal claim; the full trace format is given in Appendix~\ref{app:learning-to-go-on}.

We consider two supervision regimes: direct supervision on the input--output mapping, and trace supervision, which trains the model to imitate the intermediate state updates. For each regime, we compare standard softmax scoring with SSA, the alternative scoring function of \citet{naim:etal:2026}, that is designed to reduce softmax saturation and preserve more contextual information across tokens.

\hidden{
Following \cite{naim:etal:2026} who argue that their ssa scoring function improves performance on tasks like quantification, we tested both a standard softmax model and one with their ssa scoring function.}

\begin{figure}[h]
    \centering
    \includegraphics[width=0.95\linewidth]{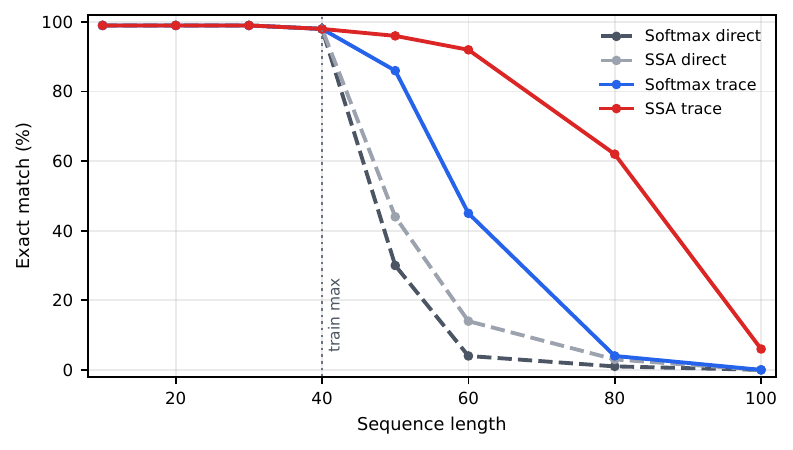}
    \caption{Direct supervision trains only on input-output mappings; trace supervision trains on state updates.}
    \label{fig:turing-exp}
\end{figure}

Figure~\ref{fig:turing-exp} shows that trace supervision moves the OOD cliff farther from the training boundary: the model applies the update rule for longer sequences than under direct final-answer supervision. However, the cliff is not eliminated. The gain comes from additional procedural structure, not from ordinary input--output use alone.

\hidden{
Figure~\ref{fig:turing-exp} shows the effect of intermediate-state supervision in this diagnostic. Training on computational the length moves the cliff significantly further from the training boundary. The model continues to apply the the update rule for longer sequences than under final-answer supervision. However, the cliff is not eliminated. Performance still degrades once the sequence length moves far enough beyond the training regime. The evidence strongly suggests that the gain comes from additional procedural structure, not from ordinary input--output use alone.}

Procedural supervision can extend the range over which a learned rule is applied, but it does not remove Kripke's point that finite traces under-determine how the procedure should extend beyond the training regime.
The experiments in this section share a shape. Each intervention supplies structure that the training distribution does not, and each extends the range over which the rule is applied correctly without fixing how it should extend further. 


\hidden{
The results show that small benchmark slices can support very different conclusions about the same model: with five examples, the mean range across seeds is 45 percentage points, and the maximum range reaches 100 points. As the subset size grows, the estimate becomes more stable. This is expected, but it also clarifies the status of benchmark scores. Larger samples give better estimates of performance on the benchmark distribution; they still do not by themselves establish that the model has acquired the rule or concept governing principled extensions of the task.

These results do not show that benchmarks are useless. They show that benchmark scores should be treated as sample-dependent estimates. For claims about semantic generalization, benchmark reporting should therefore be paired with uncertainty estimates and, where possible, rule-preserving OOD diagnostics. Per-model and per-task results are reported in Appendix~\ref{app:benchmark-variance}.
}

\section{Numerical Encoder and Decoder Implementation Details}
\label{app:numeric-implementation}

This appendix gives implementation details for the two structured numerical interfaces used in Table~\ref{tab:id-ood-tab}. The main text describes the interfaces at a high level: a column-state decoder, used mainly for smaller models, and an auxiliary supervised text method, used mainly for stronger pretrained models. Here we describe how the numerical encoder, embedding injection, output decoder, auxiliary losses, and evaluation are implemented.

\subsection{Data and evaluation protocol}
\label{app:data-eval-protocol}

All arithmetic experiments use synthetically generated addition examples with prompts of the form
\[
  a + b =
\]
where $a$ and $b$ are sampled uniformly from specified digit-length ranges. For a training range $1,\ldots,n$, in-distribution examples use operands with at most $n$ digits. Near-OOD examples use operands slightly beyond this range, and far-OOD examples extend the range further. The value of $n$ differs by model family, as reported in Table~\ref{tab:id-ood-tab}.

We evaluate by exact match on the full generated integer answer. This is an end-to-end metric: a prediction is correct only if the entire generated number matches the true integer answer. We therefore avoid padded digit-slot accuracy, which can be inflated by easy padding positions and does not faithfully measure numeric generation. For text-generation models, the generated string is parsed as an integer. For column-state models, the digit decoder output is converted directly into an integer. All reported headline results use deterministic decoding.

\subsection{Numerical encoder}
\label{app:numeric-encoder}

Both structured interfaces use a numerical encoder inspired by Fourier-style number representations. Given an operand value $x$, the encoder combines periodic features, a log-magnitude feature, and a sign feature. Abstractly, the whole-number embedding has the form
\[
\begin{aligned}
\mathrm{num}(x)
={}& W_{\mathrm{fourier}}\,\phi(x) \\
&+ \mathrm{MLP}\!\left(\log_{10}(|x|+1)\right)
+ e_{\mathrm{sign}}(x), \\[3pt]
\phi(x)
={}&
\left[
  \cos\!\left(\frac{2\pi |x|}{T_i}\right),
  \right. \\
&\qquad\left.
  \sin\!\left(\frac{2\pi |x|}{T_i}\right)
\right]_{i}.
\end{aligned}
\]
The periods $T_i$ are powers of ten over a wide decimal range, by default $10^{-3},10^{-2},\ldots,10^{12}$. These periodic features make decimal-scale information explicitly available. The log-magnitude term gives a coarse scale signal, and the sign embedding distinguishes positive and negative values. The final vector is projected to the hidden dimension of the backbone language model.

The two interfaces use this numerical embedding differently. The column-state decoder uses it as a replacement embedding for a special numerical token. The auxiliary supervised text method uses it additively, preserving the ordinary text tokens while adding numerical information as a side channel.
 
\subsection{Column-state decoder}
\label{app:column-state-decoder}

In the column-state interface, each input number is replaced in the prompt by a special numerical token. The integer value of that number is passed separately through the numerical encoder, producing one vector for the whole number. This vector is projected into the backbone embedding space and substituted for the embedding of the special numerical token before the sequence is passed through the transformer.

The output side is also modified. Instead of producing numeric answers through the standard language-model vocabulary, the model uses a dedicated number decoder conditioned on the transformer hidden state at the answer context. The decoder emits the answer least-significant digit first. 

For the column-state runs, operands are also represented as a column memory. Each operand is decomposed from right to left into decimal columns, for example
\[
372 \mapsto (2,7,3), \qquad 158 \mapsto (8,5,1).
\]
Each column memory entry contains a digit embedding, a column-position embedding, an operand-id embedding, and a sign embedding. This gives the decoder direct access to aligned operand columns, such as the ones digit of $a$ and the ones digit of $b$. The language-model backbone, numerical encoder, adapter, and number decoder are fine-tuned jointly. Thus, these results should not be assimilated to those of a frozen external calculator attached after the fact.

\subsection{Auxiliary supervised text method}
\label{app:auxiliary-text-method}

The auxiliary supervised text method keeps the ordinary language-model input and output pathway. Input numbers remain text tokens, and answers are generated with the standard language-model head. Unlike the column-state decoder, this method does not replace the text pathway. Instead, at each token position overlapping a number span with value $x$, the model adds a gated numerical embedding to the ordinary token embedding:
\[
e'_t = e_t + g \cdot \mathrm{num}(x).
\]
Here $e_t$ is the pretrained token embedding and $g$ is a learned gate. This additive design is motivated by FoNE-style \citep{zhou2026fone} numerical features: it preserves the existing BPE/text pathway while giving the model explicit numerical structure.

Training uses the standard language-model loss together with auxiliary number-oriented losses. These losses encourage hidden states associated with numerical expressions to encode structure such as magnitude, digit position, column information, answer length, and carry state. In the output-readout implementation used for the strongest auxiliary results, token-level heads predict the numerical chunk represented at a generated numeric token and the least-significant-digit-counted column of that token. Context-level heads predict global numerical information such as the answer length and carry pattern. The auxiliary heads are training-time supervision: the reported auxiliary models still generate answers through the ordinary language-model vocabulary. 

This distinction is important. The column-state decoder changes the numeric output interface, while the auxiliary supervised text method keeps the output interface textual and instead encourages the hidden states to expose numerical structure. The residue probes in Appendix~\ref{app:residue-probes} show that this auxiliary training makes sum-related residue information more linearly decodable, even when far-OOD exact-match generalization remains limited.

\subsection{Training and evaluation}
\label{app:training-eval-details}













\begin{figure*}[!h]
    \centering
    \includegraphics[width=\linewidth]{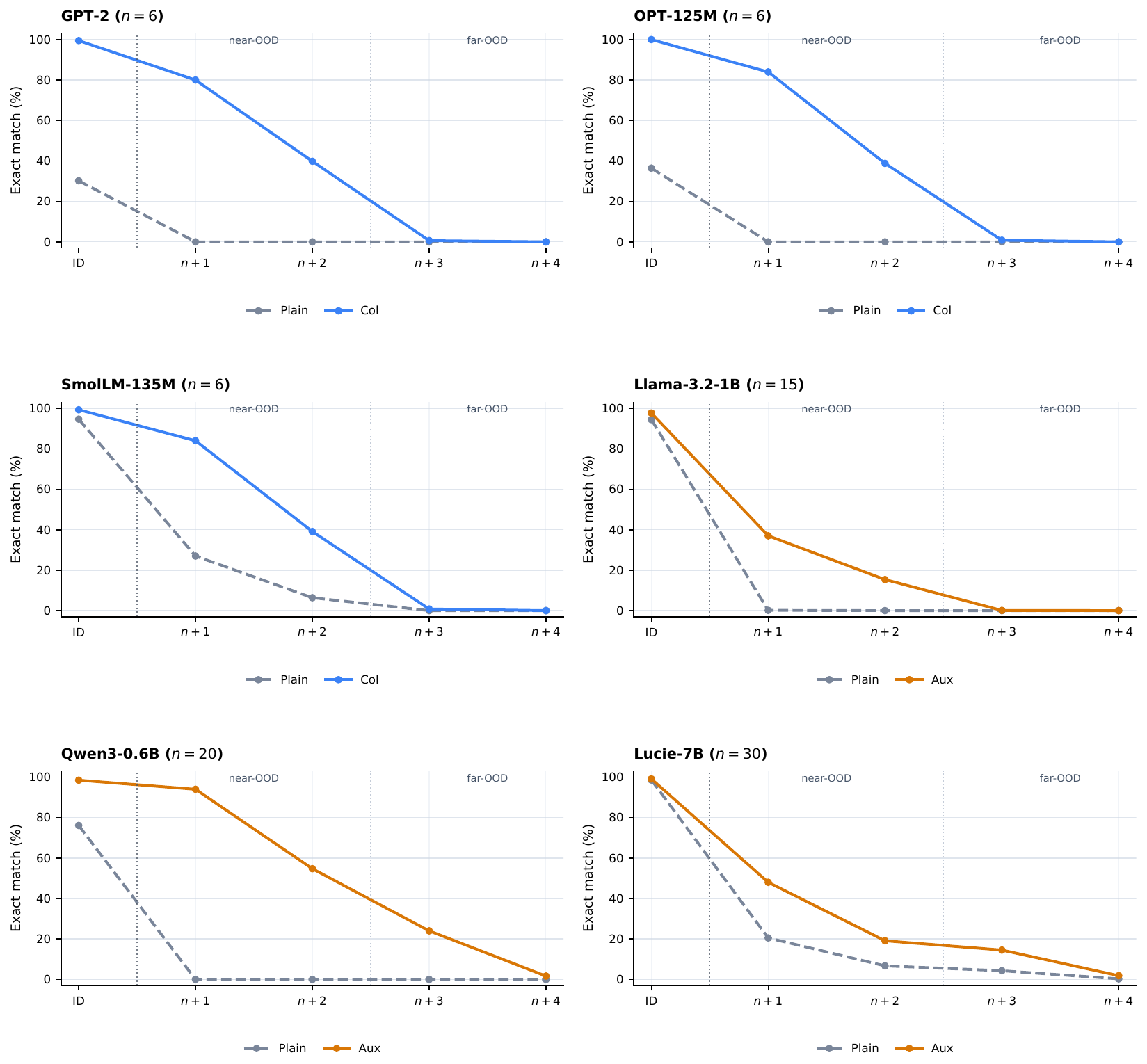}
    \caption{Per-operand-length OOD accuracy trends for addition. The x-axis shows operand length relative to the maximum training operand length: ID, $n+1$, $n+2$, $n+3$, and $n+4$. Accuracy drops sharply as operand length moves farther beyond the training regime.}
    \label{fig:add_length_breakdown}
\end{figure*}

\paragraph{Arithmetic training setup.}
All arithmetic experiments use synthetic integer expressions and report add-only evaluation. Unless otherwise stated, each run uses 50,000 training examples and 5,000 validation examples, with operands sampled uniformly from the specified digit-length range. Evaluation uses 2,000 examples per split and reports exact match on the full generated answer. Decoding is greedy and deterministic. Plain-text generations are parsed by taking the first whitespace-delimited generated token as the predicted integer; non-integer generations are counted as incorrect. 
Figure~\ref{fig:add_length_breakdown} visualizes the same trend across models and interfaces.

All headline arithmetic models are fully fine-tuned rather than trained with parameter-efficient adapters. We use AdamW with weight decay 0.01, cosine learning-rate decay, gradient clipping at 1.0, and maximum sequence length 128, except for Lucie-7B where we use maximum sequence length 192. Batch size is 32 for GPT-2, OPT-125M, SmolLM-135M, Llama-3.2-1B, and Qwen3-0.6B, and 4 for Lucie-7B. GPT-2, OPT-125M, and SmolLM-135M are trained for 30 epochs with learning rate $5\times 10^{-5}$ for plain fine-tuning; their column-state runs use learning rate $5\times 10^{-5}$ for the backbone and $10^{-3}$ for newly initialized numerical parameters. Llama-3.2-1B and Qwen3-0.6B are trained for 10 epochs; plain runs use learning rate $5\times 10^{-5}$, while auxiliary runs use backbone learning rates $5\times 10^{-5}$, with $10^{-3}$ for auxiliary parameters. Lucie-7B is trained for 10 epochs in bfloat16 with gradient checkpointing, using batch size 4 and learning rate $10^{-5}$ for plain fine-tuning; its auxiliary run uses backbone learning rate $10^{-5}$ and auxiliary-parameter learning rate $5\times 10^{-4}$. The arithmetic runs use fixed synthetic data seeds, and the reported numbers are averaged over 3 random initializations.

\begin{table*}[!h]
\centering
\small
\setlength{\tabcolsep}{4pt}
\resizebox{\linewidth}{!}{
\begin{tabular}{lcccccc}
\toprule
Model & $a\bmod 10$ & $b\bmod 10$ & $(a+b)\bmod 2$ & $(a+b)\bmod 5$ & $(a+b)\bmod 10$ & $(a+b)\bmod 100$ \\
\midrule
Llama-3.2-1B + plain FT
& 87.0\% & 98.1\% & 62.5\% & 25.4\% & 12.7\% & 1.4\% \\
Llama-3.2-1B + aux FT
& 99.6\% & 99.8\% & 100.0\% & 95.8\% & 84.8\% & 9.6\% \\
\bottomrule
\end{tabular}}
\caption{Held-out linear-probe accuracy for residue information at the answer-context hidden state. The probe asks whether base-10 arithmetic information is linearly decodable from the model state. Operand residues are already decodable after ordinary arithmetic fine-tuning, but numerical-auxiliary training makes residues of the
sum much more decodable. Chance accuracy is $1/T$ for a modulo-$T$ residue.}
\label{tab:appendix-residue-probes}
\end{table*}

\section{Residue Probes for Arithmetic Geometry}
\label{app:residue-probes}

To distinguish representational failures from generation failures, we probe whether arithmetic residue information is linearly decodable from the model hidden state at the answer context. For a period $T$, the residue $n \bmod T$ captures periodic base-10 structure: for example, $n \bmod 10$ is the final digit of $n$, and $(a+b)\bmod 10$ is the final digit of the sum.

We generate 12,000 pairs $a,b\in[0,999]$ and split them 75/25 into probe-training and held-out sets. For each pair, we collect hidden states at the last token of $a$, the last token of $b$, and the answer-context position in prompts of the form $a+b=$. We then fit ridge probes with regularization $10^{-2}$ to predict Fourier targets for residues modulo $2,5,10,$ and $100$. Held-out residue accuracy is computed by mapping the predicted Fourier vector back to the nearest residue class. Results are reported in Table~\ref{tab:appendix-residue-probes}.

\section{Algorithms}
\label{app:algo}
{\bf Quantification:} From Section~\ref{sec:limits}, we can test quantification in a setting with a linearized domain and a single property. This abstracts a common natural-language reasoning pattern: statements such as ``all emails from John before July 15 require action'' require scanning a finite sequence while preserving the same universal rule as the sequence grows. 
To test the effectiveness of abstraction and recursion, we then supply the general algorithm used in Section~\ref{sec:limits}: assume ``every number is positive'' is true until you come to a negative number. Then predict false for the rest of the autoregressive predictions.

\noindent{\bf Addition:} For two one digit numbers $u, v$ create two registers U, V with $u \in U, v \in V$.  Convert u, v into successor notation; i.e. if $u$ is 3, then set $u = 0'''$.  Now remove one $'$ from $u \in U$ and add it to $v$ in $V$; i.e. if $v = 0s''$, then after the operation you should have $v = 0'''$.  Continue this operation until u = 0, and return the result in V.

For two \(n\)-digit numbers \(x=x_1\cdots x_n\) and \(y=y_1\cdots y_n\), let \(x_j,y_j\in\{0,\ldots,9\}\), where \(j=n\) is the units column and \(j=1\) is the most significant column. Create registers \(X_j,Y_j\) for each column \(j\), initialized with \(X_j=x_j\) and \(Y_j=y_j\). Also create an extra leading register \(X_0=0\).

Starting from \(j=n\) and moving leftward to \(j=1\), add the contents of \(Y_j\) into \(X_j\). If after this operation \(X_j\ge 10\), write
\[
X_j := X_j-10
\]
and increment the next higher column:
\[
X_{j-1} := X_{j-1}+1.
\]
After all columns have been processed, return
\[
X_0 * X_1 * \cdots * X_n,
\]
omitting \(X_0\) if it is \(0\), where \(*\) denotes digit concatenation.

\section{Quantification Trace Diagnostic: Implementation Details}
\label{app:learning-to-go-on}

The task is universal quantification over signed integer sequences. Input numbers are represented by tokens of the form \texttt{P004} for positive integers and \texttt{N002} for negative integers. The target output is the truth value of the sentence ``all numbers so far are positive'' at each prefix of the sequence.

\paragraph{Data.}
Training sequences have lengths sampled uniformly from $1$ to $40$. We use integer magnitudes from $1$ to $99$. In $20\%$ of examples all numbers are positive. Otherwise, the first negative position is sampled uniformly from the sequence, all preceding numbers are positive, and later numbers may be positive or negative. Evaluation uses the same number distribution but tests longer sequence lengths: $10,20,30,40,50,60,80,100$.

\paragraph{Training formats.} We compare two target formats. In the final-answer format, the model is trained only to produce the prefix truth values:
\begin{verbatim} 
<bos> <seq> P004 P009 N002 P007 
<ans> T T F F <eos>
\end{verbatim}

In the trace format, the model is trained to produce the intermediate state update before the final answer:
\begin{verbatim}
<bos> <seq> P004 P009 N002 P007 <trace>
S T R P004 O T <sep>
S T R P009 O T <sep>
S T R N002 O F <sep>
S F R P007 O F <sep>
<ans> T T F F <eos>
\end{verbatim}
Here \texttt{S} marks the previous state, \texttt{R} the read input token, and \texttt{O} the output state after reading that token. The loss is applied only after the target marker, i.e. after \texttt{<ans>} in the final-answer condition and after \texttt{<trace>} in the trace condition.

\paragraph{Model.}  All four models are trained from scratch with the same decoder-only transformer architecture: 8 layers, hidden size 512, 8 attention
heads, MLP size 2048, sinusoidal positional embeddings, maximum sequence length 1024, and dropout 0.0. The vocabulary contains the special tokens, truth-state tokens, and signed integer tokens. We train for 4000 steps with batch size 64, AdamW, learning rate $3\times 10^{-4}$, weight decay 0.01, 200 warmup steps, and bfloat16 precision.

\section{GSM8K transfer: dataset variants and complete results}
\label{app:gsm-scaling-results}

The transfer evaluation uses 214 paired problems derived from a curated subset of GSM8K test questions. Each source problem is paired with a target problem that changes the scenario and bound quantities while preserving the reference expression. Variables denote quantity roles, rather than their order of occurrence. The same source IDs and mathematical relationships are used in all four target conditions described in Table~\ref{tab:gsm-variants}.

\begin{figure*}[h]
\centering
\includegraphics[width=0.85\textwidth]{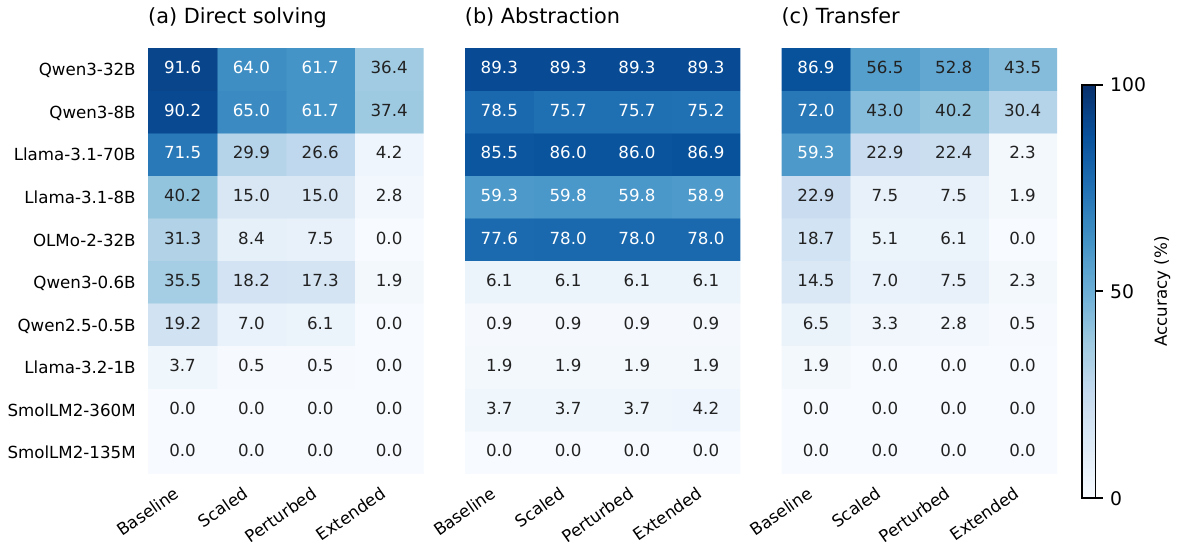}
\caption{Complete direct-solving, abstraction, and transfer results on 214 matched pairs per condition. Scaled, perturbed, and extended denote the conditions in Table~\ref{tab:gsm-variants}. Source prompts remain unchanged across columns; the scaled and perturbed conditions share generated abstractions. Values are exact numerical or algebraic-equivalence accuracies, as appropriate.}
\label{fig:gsm-all-models}
\end{figure*}

\begin{table*}[t]
\centering
\small
\renewcommand{\arraystretch}{1.2}
\begin{tabularx}{\textwidth}{@{}l l X@{}}
\toprule
Condition & Largest target input & Construction \\
\midrule
Baseline & Unscaled & Retains the target quantities assigned when the paired problems were constructed. \\
Scaled quantities & $[10^{4},10^{6})$ & Places the largest input in the five- or six-digit range while preserving structural ratios, fractions, and domain constraints. \\
Fractional perturbation & $[10^{4},10^{6})$ & Uses the scaled condition, adding 0.01--0.99 to one eligible price or measurement in 66 problems. The other 148 problems remain unchanged. \\
Extended scale & $[10^{12},10^{13})$ & Places the largest input in the 13-digit range while preserving the reference expression and domain constraints. \\
\bottomrule
\end{tabularx}
\caption{GSM8K transfer dataset variants, each containing the same 214 pairs. Ranges describe the largest bound input per target problem, not every operand or the answer. Baseline and scaled quantities can overlap; the conditions are not disjoint magnitude bins.}
\label{tab:gsm-variants}
\end{table*}

The scaled-quantity condition does not require every operand or intermediate result to be integral: structural ratios and fractions remain part of the problems. The fractional intervention affects only 66 pairs, so its aggregate score averages changed and unchanged problems. Source questions are identical across conditions. The scaled and perturbed conditions reuse the same generated abstractions; baseline and extended-scale evaluations generate them separately. Variation in abstraction scores therefore reflects separate executions on unchanged source prompts, not abstraction from larger source quantities.

We distinguish direct solving (Eval~1), source abstraction (Eval~2), and transfer (Eval~3). Direct solving receives the target question; abstraction receives the source question and quantity legend; transfer receives the generated expression and supplied target bindings in a second model call. No reference-expression fallback is supplied when abstraction fails. Expressions are scored by exact rational-function equivalence, and numerical answers by exact equality. Figure~\ref{fig:gsm-all-models} reports all ten models under these criteria. Llama, OLMo, Qwen2.5, and SmolLM2 labels denote their Instruct variants; OLMo-2-32B abbreviates OLMo-2-0325-32B-Instruct.

The application decline extends beyond the Qwen models discussed in the main text. Llama-3.1-70B-Instruct's transfer accuracy falls from 59.3\% at baseline to 2.3\% at extended scale, while its source-abstraction accuracy remains above 85\%. OLMo-2-0325-32B-Instruct retains approximately 78\% abstraction accuracy but obtains no correct transfer answers at extended scale. These results distinguish generating an appropriate expression from reliably executing it under the evaluated conditions.

\paragraph{Controls and decoding.} Transfer receives supplied variable bindings and uses two model calls. At extended scale, Qwen3-32B scores 45.3\% on a direct control with the same bindings, compared with 43.5\% for transfer and 36.4\% for unassisted direct solving. The apparent advantage over unassisted solving therefore does not isolate a benefit from generating the expression. All GSM conditions use greedy decoding, with Qwen thinking disabled, a 256-token abstraction budget, and a 1,024-token answer budget. Exact scoring does not accept rounded approximations to nonterminating fractions.

\paragraph{Scope.} The source questions form a curated rather than random sample and may have appeared in pretraining. The authored target scenarios and quantity mappings have not received independent human review. Baseline targets change both context and quantities relative to their sources, and extended-scale quantities need not describe realistic situations. The results characterize exact-answer performance under the stated construction and generation budgets; they do not establish unrestricted generalization or failure under every computational regime.

\section{Composition of symbolic functions}
\label{app:symbolic-composition}

We evaluate 50 algebraically distinct rules, ten at each operation count from one to five, using addition, subtraction, multiplication, doubling, and squaring. Each rule has 20 input and story instances, giving 1,000 problems per condition. Numerical inputs are distinct integers from 1 to 9; all intermediate values remain within $\pm10{,}000$. Symbolic counterparts replace the inputs with $a$ and $b$. Rules are explicitly supplied, so the experiment tests composition from a stated procedure rather than inference of hidden operations. Operation count refers to the stated procedure, with doubling and squaring each counted as one operation.

We compare \emph{direct prompting}, which requests only the final answer or expression; \emph{provided decomposition}, which supplies numbered operations and requests intermediate results; and \emph{model-generated decomposition}, which describes the rule in a paragraph and asks the model to determine its intermediate computations. Direct prompting uses the same numbered procedure as provided decomposition. Model-generated decomposition changes both presentation and response instructions, so this comparison does not isolate a single prompting effect. Numerical answers use exact integer equality and symbolic answers use polynomial equivalence. We report audited scores that normalize unambiguous answer formatting. Each condition uses greedy decoding, a 1,024-token output budget, and disabled Qwen thinking.

\begin{figure*}[t]
\centering
\includegraphics[width=\textwidth]{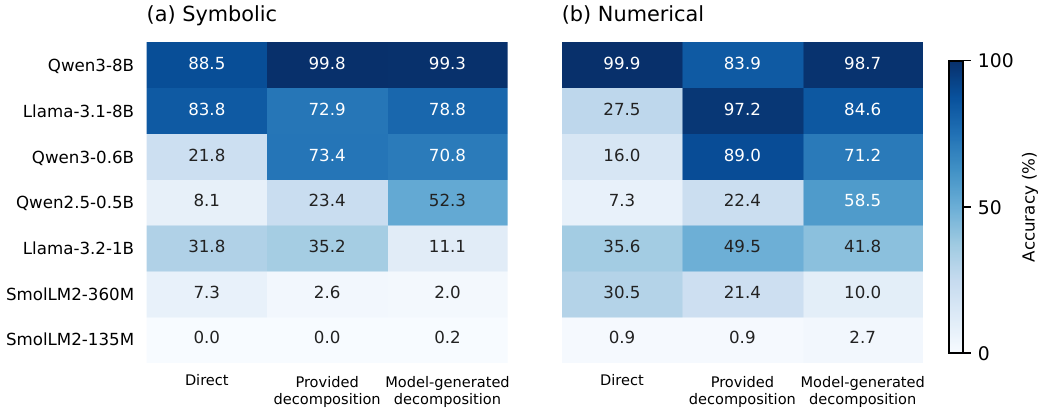}
\caption{Symbolic and numerical composition accuracy on 1,000 problems per condition. Direct prompting requests a final output; provided and model-generated decomposition request intermediate results. Scores measure presentation-normalized final-answer correctness, not correctness of every intermediate step.}
\label{fig:composition}
\end{figure*}

\begin{figure*}[htbp]
\centering
\includegraphics[width=\linewidth]{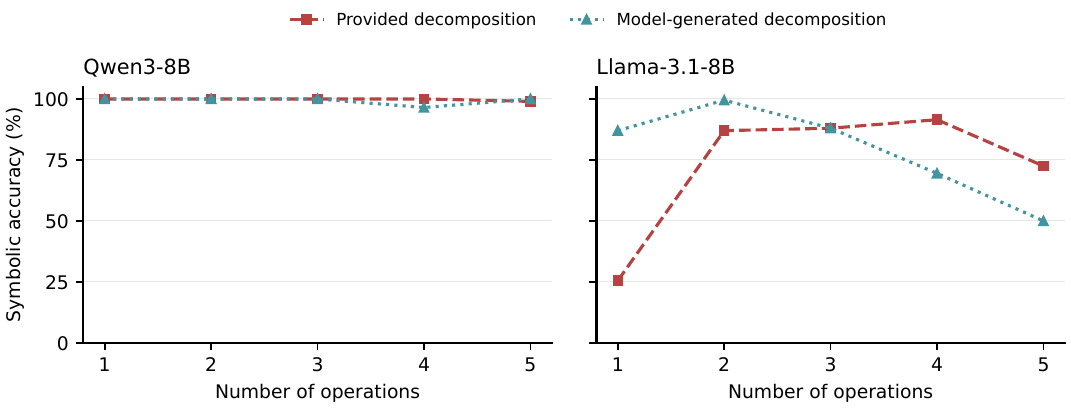}
\caption{Symbolic composition by operation count with provided and model-generated decomposition. Each point averages 200 instances from ten rules. Direct prompting is omitted. These bounded explicit-rule tests do not establish unrestricted compositional generalization.}
\label{fig:composition-complexity}
\end{figure*}

Qwen3-8B obtains 99.8\% symbolic accuracy with provided decomposition and 99.3\% with model-generated decomposition; its numerical scores are 83.9\% and 98.7\%. Llama-3.1-8B-Instruct obtains 72.9\% and 78.8\% symbolically, compared with 97.2\% and 84.6\% numerically. Under direct prompting, Llama achieves 83.8\% symbolically but only 27.5\% numerically, despite 99.6\% accuracy on the 1,029 unique isolated primitive calculations. Neither representation nor prompting condition is uniformly easier, and difficulty with composition cannot always be reduced to difficulty with the constituent calculations.

The operation-count comparison retains only provided and model-generated decomposition. At five operations, Qwen3-8B obtains 99.0\% and 100.0\% symbolic accuracy, respectively, compared with 72.5\% and 50.0\% for Llama-3.1-8B-Instruct. Performance is not uniformly monotonic in operation count. Each count contains different rules, with varying primitive choices, branching structures, and output magnitudes. The curves are therefore descriptive comparisons, not an isolated estimate of the effect of adding one operation.

\section{Complete ASDiv addition and subtraction results}
\label{app:asdiv-results}

Table~\ref{tab:asdiv-all} reports all seven models on the same 190 addition and subtraction problems. The main-text figure focuses on the two 8B models, but their original-to-abstracted gap is not universal: several smaller models match or exceed their numerical scores under five-shot prompting. Numerical answers use numeric equality, whereas abstracted answers use formula exact match rather than algebraic equivalence. The comparison therefore includes differences in both task representation and scoring criterion.

\begin{table}[H]
\centering
\small
\setlength{\tabcolsep}{3pt}
\begin{tabular}{lrrrr}
\toprule
& \multicolumn{2}{c}{Zero-shot} & \multicolumn{2}{c}{Five-shot} \\
\cmidrule(lr){2-3}\cmidrule(lr){4-5}
Model & Orig. & Abstr. & Orig. & Abstr. \\
\midrule
Qwen3-8B & 90.0 & 64.7 & 91.6 & 76.3 \\
Llama-3.1-8B & 87.9 & 56.8 & 88.4 & 63.2 \\
Qwen2.5-0.5B & 14.2 & 7.9 & 52.1 & 58.9 \\
Qwen3-0.6B & 56.3 & 24.2 & 52.6 & 52.6 \\
SmolLM2-360M & 10.5 & 0.5 & 20.0 & 45.8 \\
Llama-3.2-1B & 58.4 & 7.9 & 67.9 & 39.5 \\
SmolLM2-135M & 0.0 & 0.0 & 1.6 & 15.8 \\
\bottomrule
\end{tabular}
\caption{Complete ASDiv addition/subtraction scores (\%), with 190 problems per condition. Original scores measure numerical-answer accuracy; abstracted scores measure formula exact match. Llama, Qwen2.5, and SmolLM2 entries use Instruct variants.}
\label{tab:asdiv-all}
\end{table}


\section{Geometry rule-application experiment}
\label{app:geometry}

\begin{table}[H]
\centering
\resizebox{\linewidth}{!}{%
\begin{tabular}{lrrr}
\toprule
Model & EM & Verified EM & Unverified \\
\midrule
Llama-3.1-70B-Instruct   & 80.0 & 69.0 & 11.0 \\
Qwen3-8B                 & 66.2 & 64.8 &  1.4 \\
Qwen3-32B                & 62.0 & 61.6 &  0.4 \\
Llama-3.1-8B-Instruct    & 58.2 & 37.2 & 21.0 \\
OLMo-2-0325-32B-Instruct & 57.4 & 17.6 & 39.8 \\
Qwen3-0.6B               & 51.2 &  8.6 & 42.6 \\
Qwen2.5-0.5B-Instruct    & 22.2 &  3.4 & 18.8 \\
SmolLM2-360M-Instruct    & 21.2 &  0.0 & 21.2 \\
Llama-3.2-1B-Instruct    & 12.8 &  0.2 & 12.6 \\
SmolLM2-135M-Instruct    &  0.6 &  0.0 &  0.6 \\
\bottomrule
\end{tabular}}
\caption{Geometry results (\% of 500). EM is final-answer exact match. Verified EM and Unverified partition EM: unverified responses have a correct answer but a derivation with at least one error or no adequate derivation.}
\label{tab:geometry}
\end{table}

\paragraph{Data.} Each of the ten families has 25 matched yes/no pairs; negatives perturb one coordinate or scale one point. All labels are computed with exact rational arithmetic (Table~\ref{tab:geometry-examples}).

\begin{table}[t]
\centering\small
\begin{tabular}{@{}p{0.95\columnwidth}@{}}
\toprule
\textbf{Unit sphere (3D)} \hfill \textit{yes} \\
Do all points lie on the unit sphere? Test $x^2+y^2+z^2=1$ for every point. \\
$\left(-\tfrac{7}{27},\tfrac{2}{27},-\tfrac{26}{27}\right)$, $\left(-\tfrac{2}{7},-\tfrac{6}{35},-\tfrac{33}{35}\right)$, $\left(-\tfrac{4}{13},-\tfrac{3}{13},-\tfrac{12}{13}\right)$ \\
\midrule
\textbf{Convex hull (2D)} \hfill \textit{no} \\
Is $q$ in the convex hull of $V_1,V_2,V_3$? Solve $q=\sum_i\lambda_iV_i$, $\sum_i\lambda_i=1$; membership holds iff every $\lambda_i\ge0$. \\
$q=\left(\tfrac{46}{7},\tfrac{19}{3}\right)$; $V=(3,-4),(13,-4),(3,6)$ \\
\midrule
\textbf{Collinearity ($d=16$, sparse)} \hfill \textit{yes} \\
Are $P_1,P_2,P_3$ collinear? Set $u=P_2-P_1$, $v=P_3-P_1$; take the first $k$ with $u_k\neq0$, set $t=v_k/u_k$, and test $v_i=t\,u_i$ for every $i$. \\
$P_1=\{4{:}-3,\,5{:}3,\,6{:}3,\,12{:}7\}$, $P_2=\{4{:}-7,\,5{:}9,\,6{:}-2,\,12{:}4\}$, $P_3=\{4{:}-15,\,5{:}21,\,6{:}-12,\,12{:}-2\}$ \\
\bottomrule
\end{tabular}
\caption{Example problems. Each prompt states the rule and asks for the substitution and exact calculation before a final yes/no. Sparse vectors list nonzero coordinates only.}
\label{tab:geometry-examples}
\end{table}

\paragraph{Verification.} Every response with a correct final answer was audited. Each displayed claim was checked against the input with exact rational arithmetic, and the derivation was labeled \emph{correct}, \emph{incorrect} (at least one uncorrected error, including errors that do not change the answer), or \emph{insufficient} (no error, but no adequate derivation, e.g., a bare ``Yes''). Only \emph{correct} counts toward verified EM.

\section{Recursive addition: setup and complete results}
\label{app:recursion}

\paragraph{Data.}
We sample 700 addition problems, 100 at each operand length $n\in\{5,10,\ldots,35\}$. Both operands have exactly $n$ digits and are drawn uniformly; duplicate pairs, in either order, are rejected. Carry patterns are neither balanced nor selected. 

\paragraph{Conditions.}
Every problem appears in three conditions. \emph{Direct} asks for the sum only. \emph{CoT} asks for a brief plan and a step-by-step calculation, followed by \texttt{Answer:~<sum>}. \emph{Recursive rule} adds the procedure in Table~\ref{tab:recursion-rule} and asks the model to output every entry state, including the initial state and the base case, as one JSON object per line with keys \texttt{step}, \texttt{A}, \texttt{B}, \texttt{C}, \texttt{O}, followed by \texttt{Answer:~<sum>}. An $n$-digit problem requires $n$ recursive calls and $n+1$ recorded states. Each call uses only single-column addition, integer division, remainder and string concatenation.

\begin{table}[t]
\centering
\scriptsize
\begin{tabular}{@{}l@{}}
\toprule
\begin{minipage}{0.97\columnwidth}
\begin{verbatim}
ADD(A, B, C, O):
    record the entry state (A, B, C, O)
    if A == 0 and B == 0:
        if C == 1: O = "1" + O
        return O with leading zeros removed, or "0"
    T = (A % 10) + (B % 10) + C
    D = T % 10
    return ADD(A // 10, B // 10, T // 10, str(D) + O)

Start with ADD(A, B, 0, "").
\end{verbatim}
\end{minipage}\\
\midrule
\texttt{95 + 17}:\quad
\texttt{(95,17,0,"")} $\rightarrow$ \texttt{(9,1,1,"2")} $\rightarrow$ \texttt{(0,0,1,"12")} $\rightarrow$ \texttt{112}\\
\bottomrule
\end{tabular}
\caption{The recursive rule supplied in the prompt (abridged from the prompt's register definitions), with an example execution. $A,B$ are integer input registers, $C$ a carry bit, and $O$ a string that keeps its leading zeros until the base case.}
\label{tab:recursion-rule}
\end{table}

\begin{figure*}[h] 
\centering 
\includegraphics[width=\textwidth]{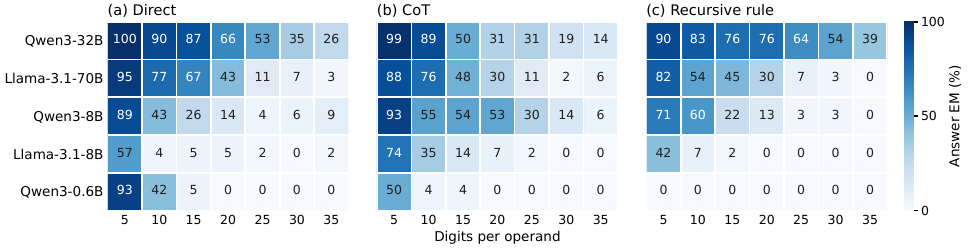} 
\caption{Answer EM (\%) for direct prompting, CoT and the supplied recursive rule on the same 100 problems per length.} 
\label{fig:recursion-all} 
\end{figure*}

\paragraph{Decoding.}
All models run in vLLM with greedy decoding, an 8,192-token context, and the same 7,168-token output budget in every condition; Qwen3 thinking is disabled. 

\paragraph{Scoring.}
EM reads the final answer line, tolerating presentation variants such as markdown or a restated ``the sum is'', and counts unparseable or truncated responses as incorrect. Verified EM additionally requires the recorded states to equal the reference sequence. We extract every JSON state, tolerating code fences, prose, extra keys, numeric or string registers, and step labels. The output register $O$ must keep its leading zeros, as the prompt requires. Comparing $O$ numerically and ignoring states after the base case still yields no correct execution. We also check each call locally, from the model's own previous state: whether the last digits of $A$ and $B$ are removed, whether the new carry equals $\lfloor T/10\rfloor$, and whether $O$ gains the digit $T \bmod 10$.

\paragraph{Results.}
Figure~\ref{fig:recursion-all} compares the three conditions at 5--35 digits, where all fifteen model--condition runs are complete. Supplying the rule does not remove the decline with length. For Qwen3-32B, the recursive prompt gives higher EM than both baselines from 20 digits on (39\% vs.\ 26\% direct and 14\% CoT at 35 digits; 28\% vs.\ 13\% direct at 50). For the other models it stays below the better baseline at every length except one (Qwen3-8B at 10 digits). At 50 digits our Llama-3.1-70B-Instruct run covers only 28 of the 100 problems (none correct), so Figure~\ref{fig:recursion} stops at 45 digits.

Qwen3-32B removes digits correctly in more than 99.9\% of calls. Its new carry, however, is correct in only 49.4\% of calls, about what a constant guess achieves (carries are 0 in 49\% of reference steps), and its output digit in only 14.2\%. In 941 of its 1,000 responses it also continues past the base case. Its answers are largely independent of its trace: only 28 of its 594 correct sums equal the value its final recorded state returns. Table~\ref{tab:recursion-example} shows a typical case.

\begin{table}[h]
\centering
\footnotesize
\setlength{\tabcolsep}{4pt}
\resizebox{0.8\columnwidth}{!}{%
\begin{tabular}{@{}rrrll@{}}
\toprule
Step & $A$ & $B$ & Ref.\ $(C,O)$ & Qwen3-32B \\
\midrule
0 & 55900 & 40560 & 0, \texttt{""} & 0, \texttt{""} \\
1 & 5590 & 4056 & 0, \texttt{"0"} & 0, \texttt{"0"} \\
2 & 559 & 405 & 0, \texttt{"60"} & 0, \texttt{"00"} \\
3 & 55 & 40 & 1, \texttt{"460"} & 0, \texttt{"000"} \\
4 & 5 & 4 & 0, \texttt{"6460"} & 0, \texttt{"0000"} \\
5 & 0 & 0 & 0, \texttt{"96460"} & 0, \texttt{"00000"} \\
6 & 0 & 0 & -- & 1, \texttt{"00000"} \\
\midrule
\multicolumn{3}{@{}l}{Answer line} & 96460 & 96460 \\
\multicolumn{3}{@{}l}{Returned by final state} & 96460 & 100000 \\
\bottomrule
\end{tabular}%
}
\caption{Qwen3-32B on $55900+40560$. It updates $A$ and $B$ correctly but prepends 0 to $O$ at every step, adds a state after the base case, and still gives the correct sum.}
\label{tab:recursion-example}
\end{table}

\end{document}